\documentclass[sigconf,nonacm]{acmart}
\AtBeginDocument{%
  }

\copyrightyear{2026}
\acmYear{2026}
\setcopyright{cc}
\setcctype{by-nc-nd}
\acmConference[BCB '26]{17th ACM International Conference on Bioinformatics, Computational Biology and Health Informatics}{June 30-July 03, 2026}{Rende (CS), Italy}
\acmBooktitle{17th ACM International Conference on Bioinformatics, Computational Biology and Health Informatics (BCB '26), June 30-July 03, 2026, Rende (CS), Italy}
\acmDOI{10.1145/3807503.3819574}
\acmISBN{979-8-4007-2653-8/2026/06}

\thanks{Accepted to the 17th ACM International Conference on Bioinformatics, Computational Biology and Health Informatics (BCB '26), 2026. Author-created version. Final Version of Record available at \url{https://doi.org/10.1145/3807503.3819574}.}

\newcommand\zw[1]{{\color{black} {#1}}}

\newcommand\fz[1]{{\color{black} {#1}}}

\usepackage{subcaption}

\usepackage{hyperref}
\hypersetup{
    colorlinks=true,
    linkcolor=blue,
    filecolor=blue,      
    urlcolor=blue,
    citecolor=blue,
}

\begin{document}


\title{AdaSM: Subgroup-Aware Survival Analysis via Mixture-of-Experts}
\title{Modeling Patient Heterogeneity in Survival Analysis through Mixture-of-Experts Representations}
\title{A Dual Mixture-of-Experts Framework for Discrete-Time Survival Prediction}
\title{Expert-Driven Survival Machines: Improving Stratification and Interpretability in Multiple Clinical Cohorts}



\author{Farica Zhuang}
\email{farica@seas.upenn.edu}
\affiliation{%
  \institution{University of Pennsylvania}
  \city{Philadelphia}
  \state{Pennsulvania}
  \country{USA}
}

\author{Zixuan Wen}
\email{zxwen@sas.upenn.edu}
\affiliation{%
  \institution{University of Pennsylvania}
  \city{Philadelphia}
  \state{Pennsulvania}
  \country{USA}
}

\author{Christos Davatzikos}
\email{christos.davatzikos@pennmedicine.upenn.edu}
\affiliation{%
  \institution{University of Pennsylvania}
  \city{Philadelphia}
  \state{Pennsulvania}
  \country{USA}
}

\author{Li Shen}
\authornote{Correspondence. lishen@pennmedicine.upenn.edu.}
\email{li.shen@pennmedicine.upenn.edu}
\affiliation{%
  \institution{University of Pennsylvania}
  \city{Philadelphia}
  \state{Pennsulvania}
  \country{USA}
}

\renewcommand{\shortauthors}{Zhuang et al.}

\begin{abstract}
  Survival prediction plays a central role for healthcare providers and clinical researchers. Accurate risk stratification enables early intervention and improved patient management. 
%
\zw{Most existing deep survival models learn one common feature representation for all patients, which may hide important differences between patient subgroups. In contrast, a Mixture-of-Experts (MoE) framework allows different parts of the model to focus on different patient patterns, leading to more individualized representations.}
%
  \zw{Therefore, in this work,} we propose a mixture-of-experts enhanced adaptive deep clustering survival framework (AdaCSM) for modeling \zw{such} heterogeneous survival patterns. We introduce a routing-based expert mechanism that enables conditional specialization within a parametric survival modeling framework. The proposed architecture allocates patients to specialized risk predictors dynamically while preserving the patient survival and subtype clustering objectives. We compare our method with state-of-the-art survival and deep clustering models on multiple real-world longitudinal clinical cohorts spanning diverse disease domains. The proposed method demonstrates improved predictive performance and leads to interpretable results in survival analysis.
\end{abstract}




\maketitle

\section{Introduction}
Time-to-event prediction (survival analysis) is designed to predict when an event is likely to happen, not just whether it happens. It has been widely used to predict disease or death in medicine and the public health domain. Classical statistical approaches include the accelerated failure time (AFT) models and the Cox proportional hazards (Cox PH). AFT models \cite{buckley1979linear, wei1992accelerated} are parametric models with the assumption that covariates either accelerate or decelerate the event time via a linear model with the log-transformation. While Cox PH \cite{cox1972regression}, a semi-parametric model, assumes the hazard rate for every instance is constant over time. Non-parametric models, such as \cite{bland1998survival}, give a stepwise survival curve that drops only when events occur.  

Recently, many deep learning methods have been proposed to improve time-to-event prediction. DeepSurv \cite{katzman2018deepsurv} is a Cox proportional hazards deep neural network, which predicts the effects of a patient’s covariates on their hazard rate. Deep survival machines (DSM) \cite{nagpal2021deep} is a fully parametric approach that models the survival function as a weighted mixture of individual survival distributions. Besides, a non-parametric model, Random Survival Forest (RSF) \cite{ishwaran2008random, ishwaran2014random} builds many survival trees on bootstrap samples, estimates the cumulative hazard function based on an ensemble of trees. Deep multi-task Gaussian process (DMGP) \cite{alaa2017deep} is used to capture complex non-linear interactions between the patients’ covariates and cause-specific survival times.

\zw{In medical applications, an important goal is to identify patient subgroups with similar survival patterns and risk profiles. Traditional unsupervised clustering methods, such as KMeans \cite{hartigan1979algorithm}, do not use time-to-event outcomes, so the resulting groups may not reflect clinically meaningful differences in prognosis. A simple alternative is to divide patients into groups using predefined thresholds, such as risk scores or survival quantiles, but these thresholds are often arbitrary and may miss more complex subgroup structures.}

\zw{Recent survival clustering methods address this problem by incorporating survival outcomes directly into subgroup discovery \cite{chapfuwa2020survival, hou2024DCSM, manduchi2021deep, jeanselme2022neural}. For example, Survival Cluster Analysis (SCA) \cite{chapfuwa2020survival} groups patients in a learned feature space using a flexible mixture model, while Deep Clustering Survival Machines (DCSM) \cite{hou2023deep, hou2024DCSM, wen2025multi} combines feature learning with parametric survival modeling so that mixture components correspond to probabilistic survival subtypes. However, these methods still generally use a single encoder to transform all patient variables into one common feature representation. When patient populations are diverse, such as in multimodal clinical data or across multiple cohorts, a single common representation may hide important differences between subgroups. A more detailed comparison is presented in Section~\ref{sec:related-work} ``Related Work''.}

\begin{figure}
  \includegraphics[width=0.45\textwidth]{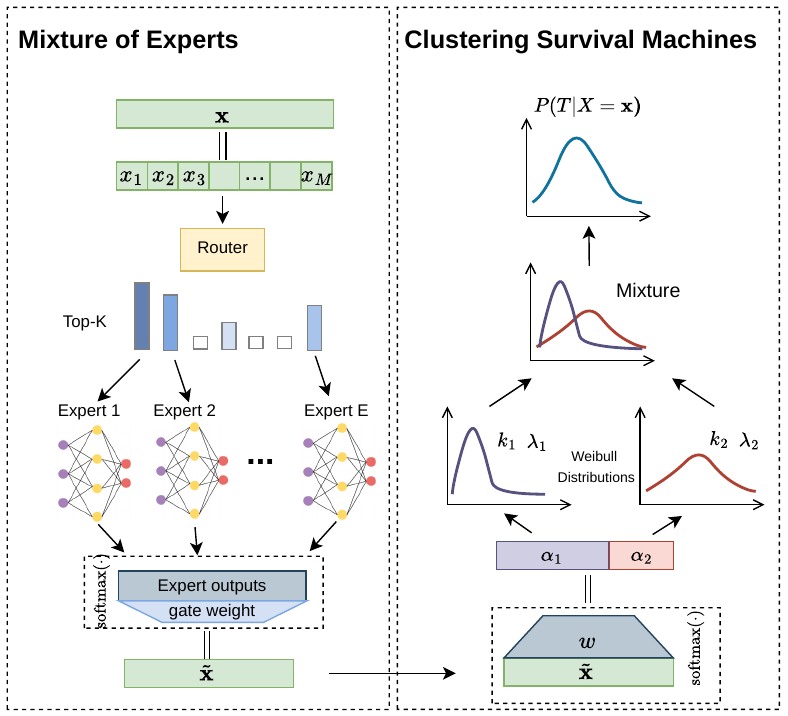}
  \caption{AdaCSM pipeline. The model uses a top‑K routing mechanism to assign each input to a subset of experts, which are subsequently processed by a Weibull‑based Deep Clustering Survival Machine. Expert outputs are combined through mixture weights produced by the router, enabling adaptive modeling of heterogeneous risk patterns.}
  \label{fig:pipeline}
  \vspace{-10pt}
\end{figure}

\zw{To address this limitation, we propose an adaptive clustering survival framework (\textbf{AdaCSM}) that captures differences across patients while preserving interpretable subtype discovery. In the AdaCSM framework (\textbf{Figure~\ref{fig:pipeline}}), a Mixture-of-Experts (MoE) encoder allows different parts of the model to focus on different types of patients. At the same time, the parametric mixture survival modeling continues to learn interpretable survival subtypes.}

\zw{Our work makes the following key contributions:
\begin{enumerate}

    \item \textbf{MoE-enhanced survival clustering architecture.} We depart from the conventional shared-encoder design by reformulating survival clustering architecture under a Mixture-of-Experts (MoE) paradigm, where representation learning is driven by adaptive expert specialization and routing. This shift enables patient-specific feature selection and individualized representations, fundamentally advancing the modeling of heterogeneous survival patterns.


    \item \textbf{Top-K sparse routing for patient-specific specialization.} We incorporate a Top-K routing mechanism so that each patient is processed by only a small subset of experts. This promotes expert specialization, improves computational efficiency, and provides a transparent view of which experts are activated for different patient profiles.

    \item \textbf{Preservation of interpretable subtype discovery.} By combining the MoE encoder with the original mixture parametric survival layer, as in deep clustering survival model, the proposed model retains the interpretable survival subtype assignments through clustering, while improving the flexibility of the learned patient representation.

    \item \textbf{Empirical validation across multiple clinical cohorts.} Experiments on several real-world clinical datasets show that the proposed framework achieves stronger subgroup separation and competitive survival prediction performance compared with existing baselines.
\end{enumerate}}

\section{Related Work}\label{sec:related-work}
Existing survival clustering methods differ in both how they encode covariates and how they define subgroups. DSM \cite{nagpal2021deep} uses a deterministic MLP to learn nonlinear representations that parameterize mixture weights and survival distributions, but this encoder is primarily optimized for individualized prediction rather than explicit subtype discovery. Accordingly, DSM does not natively perform clustering and is often used with post hoc risk stratification instead. SCA \cite{chapfuwa2020survival} employs a deterministic encoder to map covariates into a latent space for Bayesian nonparametric modeling, but it does not explicitly capture uncertainty in the learned representation. Its clustering is performed through a flexible latent mixture model, which can adaptively discover subgroups but may also produce overly fragmented subgroups that are difficult to interpret. VaDeSC \cite{manduchi2021deep, qiu2025deep} adopts a variational generative encoder with a Gaussian mixture latent prior to jointly model covariates and survival outcomes, but its learned representation can be sensitive to misspecification of the assumed feature distribution. Because its clustering is tightly coupled to that generative prior, subgroup discovery may degrade when the data deviates from the assumed model.

In contrast, \zw{Neural Survival Clustering (NSC)} \cite{jeanselme2022neural} employs an MLP-based assignment network, but this encoder serves only to assign patients to clusters rather than to model individual variation within clusters. It improves flexibility and interpretability by clustering patients via assignment to population-level neural survival components, yet potentially sacrificing individual-level covariate dependence. DCSM \cite{hou2023deep, hou2024DCSM, noshin2025integrating, wen2025multi} takes a more interpretable ground by learning a compact discriminative embedding with an MLP and using it to estimate mixture weights over fixed parametric survival experts. This design yields more clinically meaningful subgroup assignments because subjects are clustered according to weights over constant expert distributions, but it still relies on a single shared encoder that may blur subgroup-specific feature patterns. These limitations suggest that, while prior methods differ in their balance between flexibility and interpretability, most still rely on a single global covariate encoder that must explain heterogeneous patient populations with shared parameters.

Motivated by this gap, we introduce an MoE-based encoder enhanced clustering survival framework to enable subgroup specialized representation learning before survival clustering. Instead of forcing one shared MLP to encode all patients, the MoE encoder allows different experts to focus on distinct regions of the covariate space, while a gating network adaptively combines expert outputs for each individual. This design preserves the interpretable clustering mechanism while improving the encoder’s ability to capture heterogeneous, subgroup-specific feature patterns. In this way, our approach addresses a key limitation of existing survival clustering methods, the lack of encoder specialization for complex population heterogeneity.

\section{Method}
In this section, we propose the \textbf{AdaCSM} model, a mixture-of-experts enhanced adaptive deep clustering survival framework, which enables subgroup specialized representation learning before survival clustering.
Figure~\ref{fig:pipeline} illustrates the overall framework of our method. For an input vector $x$, the latent representation $\tilde{x}$ is generated by a mixture-of-experts (MoE) encoder, where a gating network adaptively assigns subjects to different experts. The survival density function is modeled as a weighted mixture of Weibull distributions, where $\alpha$ denotes the cluster-specific mixing weights. The MoE and Weibull parameters are learned jointly in an end-to-end manner from both the covariates and the survival information.


\subsection{Problem Formulation}

Let $\mathcal{D}=\{(\mathbf{x}_i, t_i, \delta_i)\}_{i=1}^{N}$ denote a dataset of $N$ individuals, where $\mathbf{x}_i \in \mathbb{R}^{d}$ represents the covariate vector describing subject $i$, $t_i$ denotes the observed follow-up time, and $\delta_i \in \{0,1\}$ is the event indicator. Specifically, $\delta_i = 1$ indicates that the event of interest (e.g., disease progression or death) was observed at time $t_i$, while $\delta_i = 0$ indicates right-censored observations where the event had not occurred by the end of follow-up.

The objective of survival analysis is to estimate the conditional time-to-event distribution given the subject's covariates. This distribution is typically characterized by the survival function
\begin{equation}
S(t \mid \mathbf{x}) = \mathbb{P}(T > t \mid X = \mathbf{x}),
\end{equation}
which represents the probability that the event time $T$ exceeds time $t$. An equivalent representation is the hazard function.

\begin{equation}
h(t \mid \mathbf{x}) = \lim_{\Delta t \to 0} 
\frac{\mathbb{P}(t \leq T < t + \Delta t \mid T \geq t, X = \mathbf{x})}{\Delta t},
\end{equation}
which captures the instantaneous risk of event occurrence at time $t$ given survival up to that time.

In many clinical settings, patient populations are heterogeneous and may exhibit multiple latent disease progression patterns. To capture this heterogeneity, we assume that the conditional survival distribution can be modeled as a mixture of $J$ latent components (or subtypes). Each component corresponds to a specialized risk predictor representing a distinct survival trajectory. Formally, the survival distribution is modeled as
\begin{equation}
S(t \mid \mathbf{x}) = \sum_{j=1}^{J} \alpha_j(\mathbf{x}) S_j(t) =  \sum_{j=1}^{J} p(j \mid \mathbf{x}) S_j(t),
\end{equation}
where $\alpha_j$ denotes the probability of assigning patient $i$ to subtype $j$, and $S_j(t )$ denotes the parametric survival function of subtype $j$. The subtype assignment probability satisfies $\sum_{j=1}^{J} \alpha_j = 1$.

The goal of the proposed framework is therefore twofold:

\begin{enumerate}
    \item \textbf{Survival prediction}: accurately estimate individualized survival distributions $S(t \mid \mathbf{x})$.
    \item \textbf{Subtype discovery}: identify latent patient subgroups characterized by distinct survival trajectories.
\end{enumerate}

\subsection{Model Overview}
The model first encodes input covariates using a Mixture-of-Experts module, where a router assigns top-K weights to multiple expert networks and aggregates their outputs into the representation $\tilde{\mathbf{x}}$. This representation is then passed to a clustering survival machine head, which predicts mixture weights over fixed Weibull distributions. The weighted combination of these survival components yields the final survival distribution and provides an interpretable subgroup assignment through the dominant component. 


\subsection{Survival Modeling}
The survival model $S(t \mid \mathbf{x})$ of subject with covariate $\mathbf{x}$ is defined as
$
S(t \mid \mathbf{x}) = \sum_{j=1}^{J} \alpha_j(\mathbf{x}) S_j(t).
$
We predefine the number of subtypes to be 2 (i.e., $J=2$), corresponding to a straightforward partition of subjects into high-risk and low-risk groups. For each subtype, the survival function $S_j(t)$ is modeled using a Weibull distribution, which is widely adopted in survival analysis due to its flexibility and its closed-form expressions for the probability density and cumulative distribution functions. Specifically,
\begin{equation}
f(t)=\frac{k}{\lambda}\left(\frac{t}{\lambda}\right)^{k-1} \exp \left({-\left(\frac{t}{\lambda}\right)^k}\right), \quad
F(t)=1-\exp\left({-\left(\frac{t}{\lambda}\right)^k}\right),
\end{equation}
where $k$ and $\lambda$ denote the shape and scale parameters, respectively. Accordingly, the survival function is given by
\begin{equation}
S(t)=1-F(t)=\exp\left({-\left(\frac{t}{\lambda}\right)^k}\right).
\end{equation}
Thus, the survival function is explicitly modeled as a weighted mixture of subtype-specific Weibull survival functions:
\begin{equation}
S(t \mid \mathbf{x}) = \sum_{j=1}^{2} \alpha_j(\mathbf{x}) S_j(t) = \sum_{j=1}^{2} \alpha_j(\mathbf{x}) \exp \left({-\left(\frac{t} {\lambda_j}\right)^{k_j}}\right),
\end{equation}
where 
\begin{equation}
\alpha_j(\mathbf{x})
=
\frac{\exp\left(w_j^T\tilde{\mathbf{x}}\right)}
{\sum_{j=1}^{2}\exp\left(w_j^T \tilde{\mathbf{x}}\right)}.
\end{equation}
Here, $\alpha_j(\mathbf{x})$ denotes the mixture weight of subtype $j$ for the subject, and can be interpreted as the probability of assignment that the subject belongs to the $j$-th survival subtype. In our framework, these subtype weights are estimated from the MoE-enhanced representation $\tilde{\mathbf{x}}$, and are used to combine the subtype-specific Weibull survival functions into an individualized survival prediction.


\subsection{Adaptive Gating Mechanism}
To better capture heterogeneous covariate patterns across subjects, we replace the first layer of the standard multilayer perceptron encoder with a Mixture-of-Experts (MoE) layer to learn $\tilde{\mathbf{x}}$. Given an input covariate vector $\mathbf{x} \in \mathbb{R}^d$, the MoE layer consists of $E$ expert networks and a gating network that adaptively assigns weights to the experts. The output representation is defined as
$
\mathbf{h}(\mathbf{x}) = \sum_{e=1}^{E} g_e(\mathbf{x}) f_e(\mathbf{x}),
$
where $f_e(\mathbf{x})$ denotes the output of the $e$-th expert and $g_e(\mathbf{x})$ is its corresponding gating weight.

In our implementation, each expert is a lightweight neural transformation composed of a linear layer followed by a nonlinearity, i.e.,
$
f_e(\mathbf{x}) = \sigma\left(\mathbf{W}_e \mathbf{x}\right),
$
where $\mathbf{W}_e$ is the learnable weight matrix for expert $e$, and $\sigma(\cdot)$ denotes the activation function. In this work, we use $\mathrm{ReLU6}(\cdot)$ as the activation. Thus, each expert learns a distinct transformation of the input covariates, allowing different experts to specialize in different regions of the feature space.

The gating network maps the input $\mathbf{x}$ to a set of routing scores, which are then normalized by a softmax function:
\begin{equation}
g_e(\mathbf{x}) = \frac{\exp \left( z_e(\mathbf{x}) / \tau \right)}
{\sum_{e'=1}^{E} \exp \left( z_{e'}(\mathbf{x}) / \tau \right)},
\end{equation}
where $z_e(\mathbf{x})$ is the routing logit for expert $e$, and $\tau > 0$ is a temperature parameter controlling the sharpness of the routing distribution. A smaller $\tau$ produces more selective routing, while a larger $\tau$ yields smoother expert combinations.

To further encourage expert specialization, we optionally apply a top-$K$ routing strategy. Specifically, only the $K$ experts with the largest gating weights are retained, while the remaining weights are set to zero. The retained weights are then re-normalized:
\begin{equation}
\tilde{g}_e(\mathbf{x}) =
\begin{cases}
\dfrac{g_e(\mathbf{x})}{\sum\limits_{e' \in \mathcal{T}(\mathbf{x})} g_{e'}(\mathbf{x})}, & e \in \mathcal{T}(\mathbf{x}),\\[8pt]
0, & \text{otherwise},
\end{cases}
\end{equation}
where $\mathcal{T}(\mathbf{x})$ denotes the set of top-$K$ selected experts for subject $\mathbf{x}$. The final MoE representation is then computed as
\begin{equation}
\mathbf{h}(\mathbf{x}) = \sum_{e=1}^{E} \tilde{g}_e(\mathbf{x}) f_e(\mathbf{x}).
\end{equation}

Here, $\mathbf{h}(\mathbf{x})$ is the intermediate output of the MoE layer, whereas $\tilde{\mathbf{x}}$ denotes the final encoder output used by the survival clustering head. This adaptive gating mechanism enables the encoder to move beyond a single shared transformation for all subjects. Instead, each subject is represented by a weighted combination of specialized experts, allowing the model to capture subgroup-specific covariate patterns before downstream survival modeling. The resulting MoE-enhanced representation is then passed to the subsequent encoder layers and ultimately used to estimate the mixture weights over subtype-specific survival experts.

\subsection{Joint Objective Function}
The proposed model is trained end-to-end by jointly optimizing the adaptive gating mechanism and the survival mixture model. For each subject, let $t_i$ denote the observed time and let $\delta_i \in \{0,1\}$ denote the event indicator, where $\delta_i=1$ indicates an observed event and $\delta_i=0$ indicates right censoring. Given the mixture-based survival model, the corresponding probability density function is defined as
$
f(t \mid \mathbf{x}) = \sum_{j=1}^{J} \alpha_j(\mathbf{x}) f_j(t),
$
and the survival function is
$
S(t \mid \mathbf{x}) = \sum_{j=1}^{J} \alpha_j(\mathbf{x}) S_j(t ),
$
where $f_j(t)$ and $S_j(t)$ denote the subtype-specific Weibull density and survival functions, respectively.

Accordingly, the likelihood contribution of subject $i$ is written as
\begin{equation}
\mathcal{L}_i
=
\left[f(t_i)\right]^{\delta_i}
\left[S(t_i)\right]^{1-\delta_i}.
\end{equation}
The overall objective is obtained by minimizing the negative log-likelihood:
\begin{equation}
\mathcal{L}_{\mathrm{surv}}
=
-\sum_{i=1}^{N}
\left[
\delta_i \log f(t_i \mid \mathbf{x}_i)
+
(1-\delta_i)\log S(t_i \mid \mathbf{x}_i)
\right].
\end{equation}

Substituting the mixture formulation into the above objective yields
\begin{equation}
\begin{aligned}
\mathcal{L}_{\mathrm{surv}}
=
-\sum_{i=1}^{N}
\Big[
&\delta_i \log \Big( \sum_{j=1}^{J} \alpha_j(\mathbf{x}_i) f_j(t_i ) \Big) \\
&+
(1-\delta_i)\log \Big( \sum_{j=1}^{J} \alpha_j(\mathbf{x}_i) S_j(t_i) \Big)
\Big].
\end{aligned}
\end{equation}

Since both the subtype mixture weights $\alpha_j(\mathbf{x})$ and the MoE representation $\tilde{\mathbf{x}}$ are differentiable functions of the input covariates, this objective can be optimized jointly with respect to all model parameters. In this way, the router, expert networks, and subtype-specific survival components are learned simultaneously in a unified framework.

\subsection{Training and Optimization}
We adopt a consistent evaluation pipeline across all methods. For time-to-event prediction, each dataset is split using a fixed random seed into 70\% training and 30\% testing sets, with the training portion further divided into training and validation subsets (60\%-10\%-30\% overall). Continuous features are normalized to [0,1] using statistics computed only from the training data, and categorical features are one-hot encoded (Appendix A.1). Hyperparameters are selected with Optuna \cite{akiba2019optuna} by maximizing validation C-Index, and the best configuration is evaluated on the held-out test set. The best hyperparameters are shown in Appendix (Appendix A.2, Appendix Tables 1-5). The experiment is repeated across five random seeds, and we report the mean and standard deviation of the C-Index and LogRank statistic.

\section{Experiment Setup}
In this section, we describe the datasets, metrics, and methods used to evaluate and benchmark our proposed framework. Our code is available at: \url{https://github.com/PennShenLab/AdaCSM}.

\subsection{Datasets}
We prepare four real-world survival datasets for experiments, and their data statistics are shown in \textbf{Table~\ref{tab:datasets}}. All datasets are analyzed under standard right-censoring, where individuals without an observed event are censored at their last follow-up time.



\begin{itemize}
    
    \item \textbf{SUPPORT \cite{knaus1995support}:} The Study to Understand Prognoses and Preferences for Outcomes and Risks of Treatments (SUPPORT) dataset sourced from a study conducted by Vanderbilt University that contains comprehensive clinical and demographic data from hospitalized adults with serious illnesses. The survival endpoint is defined as the time from hospital admission to all-cause mortality.
    
    \item \textbf{PBC \cite{fleming2013PBC}:} The Primary Biliary Cholangitis (PBC) dataset contains longitudinal clinical measurements from patients enrolled in a Mayo Clinic study of liver disease progression. We use the time-dependent version of the cohort (pbc2), which includes repeated laboratory and clinical assessments collected over follow-up. The event is defined as mortality.
    
    \item \textbf{Framingham \cite{dawber1951framingham}:} This dataset is derived from the well-known Framingham Heart Study that follows participants longitudinally to model the time to the first major cardiovascular event, including incident coronary heart disease and stroke. It incorporates baseline demographic and clinical risk factors such as age, blood pressure, cholesterol levels, and smoking history. \zw{We consider death as the event, with survival time measured from the baseline exam to death or the end of follow-up.}
    
    \item \textbf{FLCHAIN \cite{dispenzieri2012FLCHAIN}:} This dataset originates from a population-based study in Olmsted County, Minnesota, designed to evaluate the prognostic value of serum free light chain (FLC) levels. The survival endpoint is time to death from any cause. Beyond standard demographics like age and sex, the dataset includes various immunological and biochemical markers for plasma cell disorders and renal function.
\end{itemize}

A detailed description of the data preprocessing pipeline is provided in Appendix A.1.


\begin{table}[h]
\centering
\small 
\caption{\fz{Dataset statistics: including sample size ($n$), feature dimensionality after preprocessing/encoding ($d$), event rate, censoring rate, and median follow-up time.}}
\label{tab:datasets}
\begin{tabular}{l@{\hskip 0.1in}c@{\hskip 0.1in}c@{\hskip 0.1in}c@{\hskip 0.1in}c@{\hskip 0.1in}c}
\hline
\textbf{Data set} & \textbf{$n$} & \textbf{$d$} & \textbf{Event \%} & \textbf{Cens. \%} & \textbf{{\large $t$}$_{\text{\tiny \,follow-up}}$} \\ \hline
SUPPORT           & 9,105        & 59         & 68.1\%            & 31.9\%            & 233 days          \\
PBC & 1945 & 25 & 37.28\% & 62.72\% & 4.6 years         \\
FRAMINGHAM & 11,627 & 
18 & 30.3\% & 69.7 \% & 17.9 years   \\
FLCHAIN & 6,524 & 27 & 30.1\% & 69.9\% & 11.8 \textit{years} \\          \hline
\end{tabular}
\end{table}

\subsection{Evaluation Metrics} 

In the experiments, we denote the predicted risk scores as  $\hat{\eta}_i$ and the ground-truth time-to-event as $T_i$. With this, we assess the patient prognostic and subtype clustering precision by adopting the following metrics for evaluation:

The \textbf{Concordance Index (C-Index)} measures the probability that for a randomly selected pair of patients, the model predicts a higher risk for the individual who experiences the event first:

\begin{equation} 
    \text{C-Index} = \frac{\sum_{i,j \in P} I(T_i < T_j) \cdot I(\hat{\eta}_i > \hat{\eta}_j)}{\sum_{i,j \in P} I(T_i < T_j)} 
\end{equation}

where $P$ is the set of all comparable pairs in the dataset and $I(\cdot)$ is the indicator function. 

The \textbf{Log-Rank Statistic} ($\chi^2$) is used to quantify the degree of separation between the survival distributions of the latent subtypes discovered by the adaptive gating network. The magnitude of the Log-Rank statistic reflects the strength of the evidence against the null hypothesis that the subtypes share a common survival trajectory. The statistic is calculated as:
\begin{equation}
\chi^2 = \frac{(\sum_{t=1}^f O_{1t} - E_{1t})^2}{\sum_{t=1}^f V_{1t}}
\end{equation}
where $O_{1t}$ and $E_{1t}$ are the observed and expected number of events at time $t$, and $V_{1t}$ is the variance. Higher values of the $\chi^2$ statistic indicate that the model has identified subtypes with more distinct and non-overlapping clinical outcomes.

\subsection{Comparison Methods}
To evaluate the performance of AdaCSM, we compare it with several representative classical and state-of-the-art survival analysis methods. 


\begin{itemize}
    \item \textbf{Cox Proportional Hazards (Cox PH) \cite{cox1972regression}}: The Cox PH model is a widely used classical survival modeling approach based on a semi-parametric regression. It models the hazard function, which represents the instantaneous risk of experiencing the event at a given time, conditional on surviving up to that time. The model assumes proportional hazards, meaning that the hazard ratio between two individuals remains constant over time. The model parameters are estimated by maximizing the partial likelihood.


    
    
    \item \textbf{DeepSurv \cite{katzman2018deepsurv}}: DeepSurv, also known as Deep Cox, is a deep learning-based extension of the Cox PH model that replaces the linear functional with a deep neural network to capture non-linear feature dependencies. The model is trained by maximizing the Cox partial likelihood and retains the proportional hazards assumption, meaning that the relative risk between individuals is similarly assumed constant over time.
    

    \item \textbf{Deep Survival Machines (DSM) \cite{nagpal2021deep}}: DSM is a generative, fully parametric deep survival model that represents the survival distribution as a mixture of predefined parametric components (e.g., Weibull or Log-Normal distributions). Both the mixture weights and distribution parameters are predicted by a neural network conditioned on patient covariates. Unlike the Cox PH constant proportional hazards assumption, this formulation allows flexible modeling of heterogeneous survival patterns while maintaining a likelihood-based training objective.
    
    
    

    \item \textbf{Deep Clustering Survival Machines (DCSM) \cite{hou2024DCSM}}: DCSM combines discriminative feature learning with generative survival modeling to identify latent patient subgroups. This framework integrates survival modeling with representation learning to induce survival-informed clustering, where mixture components correspond to soft subtype assignments. The model is trained end-to-end using a shared monolithic neural network that simultaneously determines subtype probabilities and survival parameters for the population.

    
    
    
\end{itemize}

Consistent with previous benchmarking protocols, we ensure that all deep learning-based baselines (DeepSurv, DSM, and DCSM) are optimized using Optuna hyperparameter tuning to ensure a fair comparison with our proposed AdaCSM. 

Using two subtypes ensured a fair and uniform evaluation setting across baselines. In addition, a two-group setting aligns with the common clinical interpretation of low-risk versus high-risk stratification.

For DCSM and AdaCSM, subgroup assignments were obtained directly from the learned mixture components by assigning each patient to the component with the highest mixture weight. For methods that do not natively produce subgroup assignments (Cox PH, DeepSurv, and DSM), we derived two risk groups by splitting patients at the median predicted risk score. Patients above the median were assigned to the high-risk group, and those below the median were assigned to the low-risk group. This post hoc stratification was used only for LogRank evaluation, allowing all methods to be compared under the same two-group setting.

\section{Results}

\begin{figure}
  \includegraphics[width=0.48\textwidth]{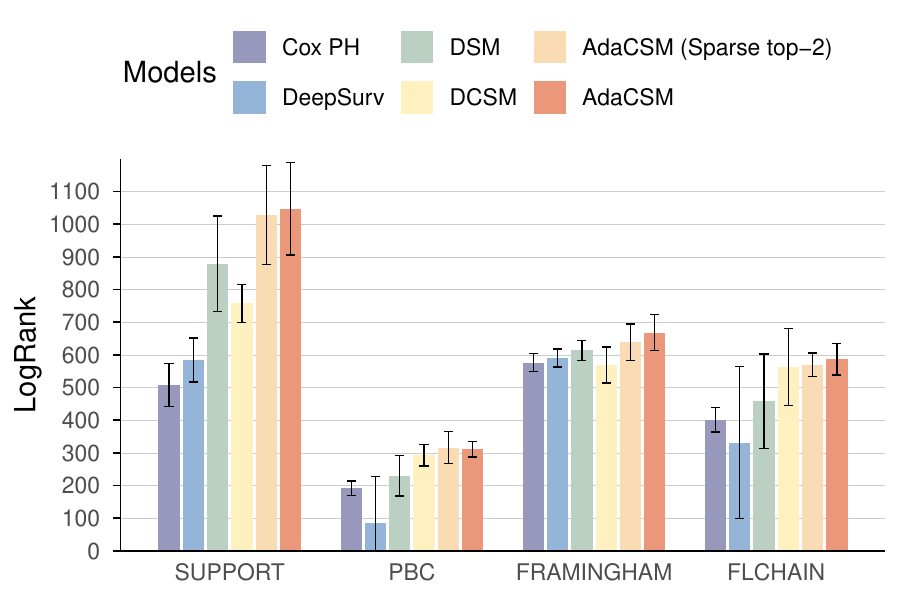}
  \caption{\zw{Comparison of LogRank statistics for clinical datasets across six survival-clustering methods: Cox PH, DeepSurv, DSM, DCSM, AdaCSM (Sparse top-2), and AdaCSM. Error bars indicate 95$\%$ confidence interval (CI).}}
  \label{fig:logrank-plot}
\end{figure}

\begin{figure*}[t]
\centering

\includegraphics[width=1\textwidth]{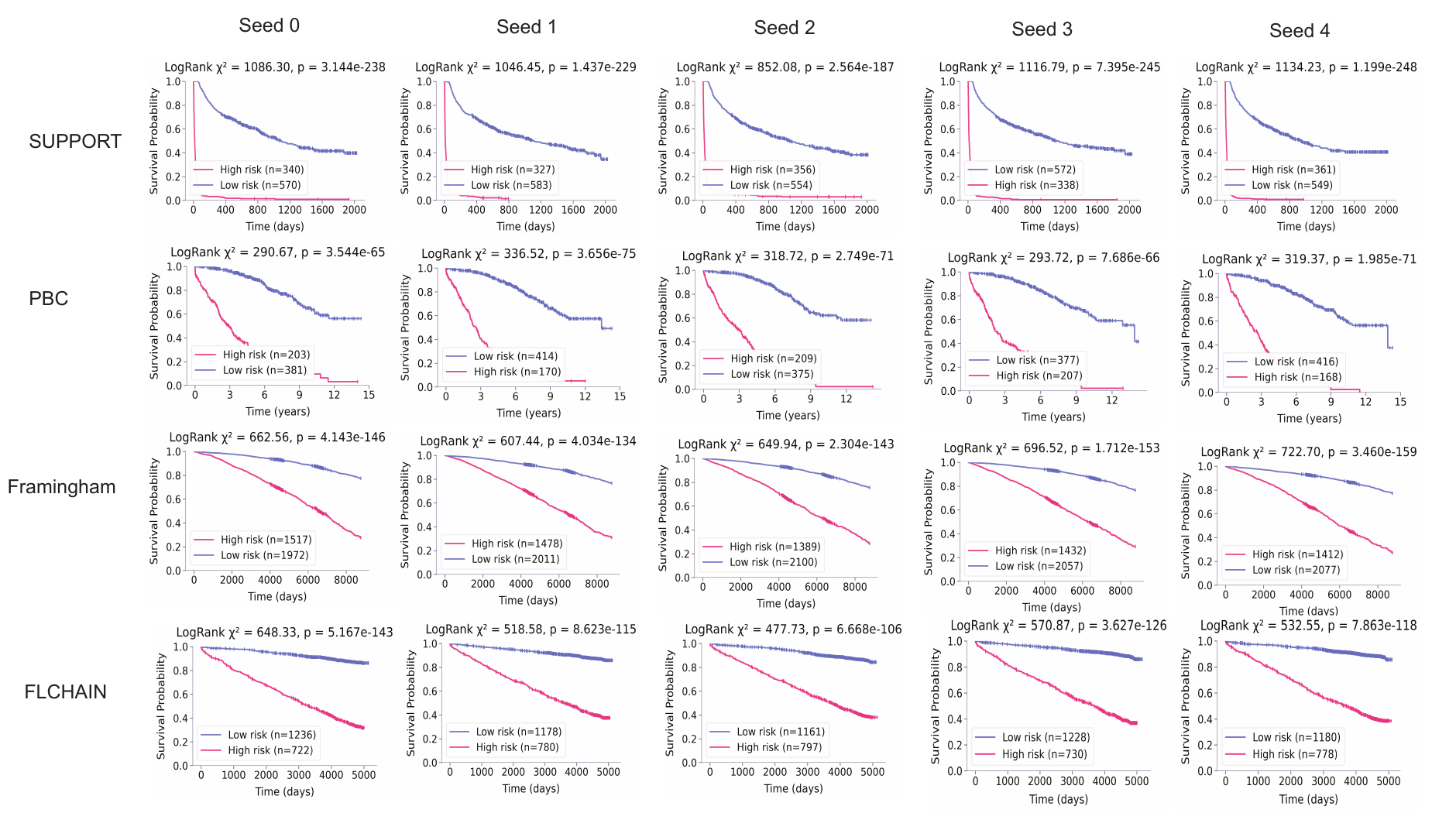}
\caption{Kaplan-Meier survival curves for the discovered subtypes across 5 random seeds for each clinical survival dataset by AdaCSM. Columns correspond to different random seeds, and rows correspond to datasets. All curves are generated using the test split.}

\label{fig:km_seeds_clinical}
\end{figure*}
\subsection{Strong Subtype Identification}

We first evaluate whether AdaCSM can identify clinically meaningful subtypes with distinct survival trajectories. As shown in 
\zw{\textbf{Figure~\ref{fig:logrank-plot}}}, AdaCSM achieves the highest LogRank statistic on all four clinical cohorts, including SUPPORT (1047.17 $\pm$ 114.04), PBC (311.80 $\pm$ 19.30), Framingham (667.83 $\pm$ 44.24), and FLCHAIN (586.56 $\pm$ 38.39). These gains are substantial relative to Cox PH, DeepSurv, DSM, and DCSM, indicating that the proposed model produces more clearly separated survival subgroups. In particular, the improvement over DCSM is consistent across all datasets, suggesting that adaptive expert specialization enhances subtype discovery beyond the original clustering survival formulation.

Subtype identification is also visually supported by the Kaplan-Meier curves in Figure~\ref{fig:km_seeds_clinical}. The red and blue Kaplan–Meier curves are visibly separated, indicating that the discovered groups correspond to distinct survival trajectories rather than arbitrary partitions. This implies AdaCSM is able to consistently identify two survival subtypes with clearly different prognoses. 


\begin{figure}
  \includegraphics[width=0.48\textwidth]{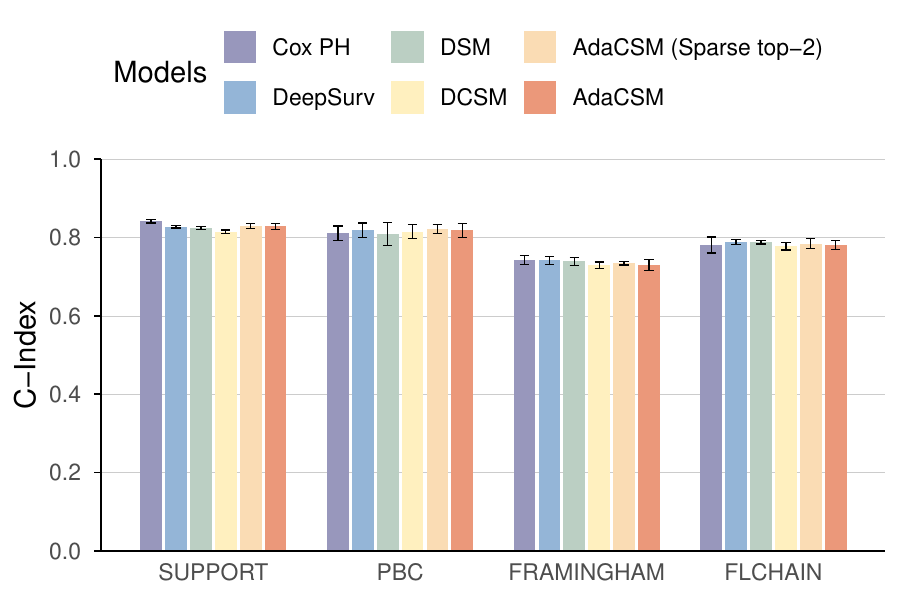}
  \caption{\zw{Comparison of C-Index statistics for clinical datasets across six survival-clustering methods: Cox PH, DeepSurv, DSM, DCSM, AdaCSM (Sparse top-2), and AdaCSM. Error bars indicate 95$\%$ confidence interval (CI).}}
  \label{fig:cindex-plot}
\end{figure}









\begin{figure}
  \includegraphics[width=0.45\textwidth]{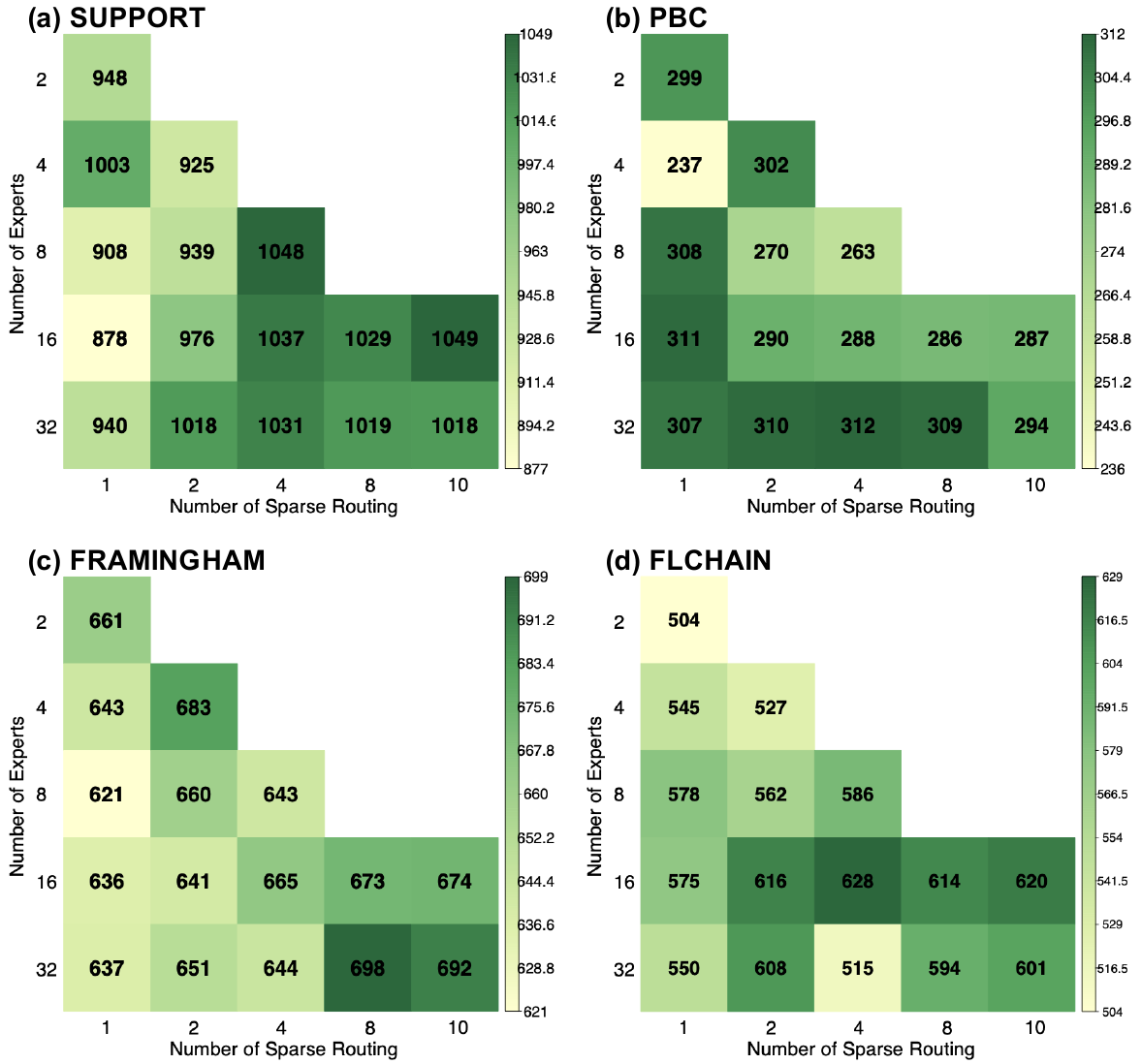}
  \caption{\zw{Sensitivity analysis for AdaCSM measured by LogRank statistic ($\uparrow$).}}
  \label{fig:sensitivity-logrank-plot}
\end{figure}

\begin{figure}
  \includegraphics[width=0.45\textwidth]{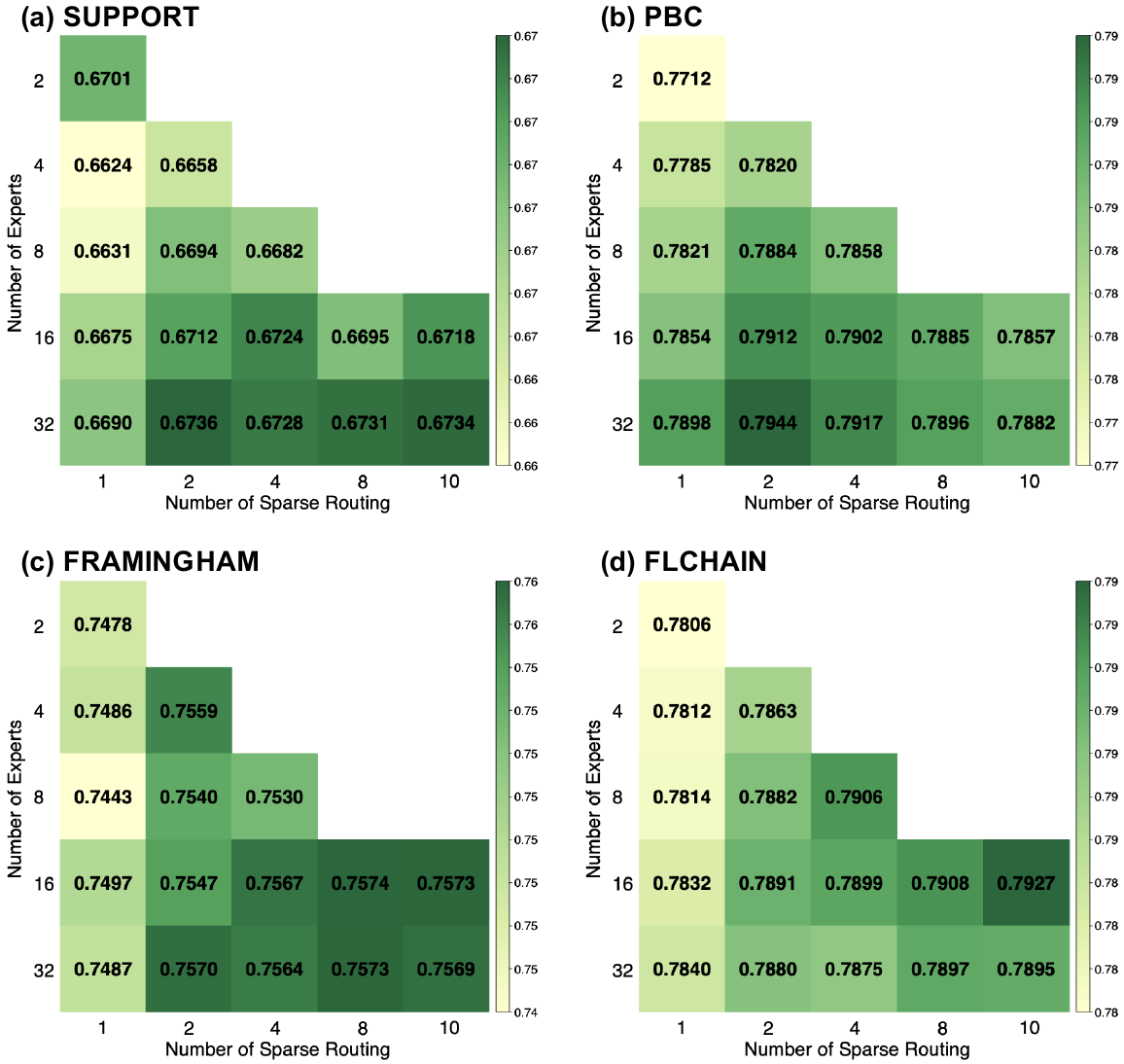}
  \caption{\zw{Sensitivity analysis for AdaCSM measured by mean C-Index statistic ($\uparrow$).}}
  \label{fig:sensitivity-cindex-plot}
\end{figure}

\subsection{Competitive Prognosis Performance}

Although AdaCSM is designed primarily for subtype discovery, we also evaluate its prognostic accuracy using the C-Index. As shown in
\textbf{Figure~\ref{fig:cindex-plot}}, AdaCSM remains competitive across all datasets, achieving 0.8284 $\pm$ 0.0060 on SUPPORT, 0.8181 $\pm$ 0.0142 on PBC, 0.7300 $\pm$ 0.0108 on Framingham, and 0.7806 $\pm$ 0.0088 on FLCHAIN. While AdaCSM does not uniformly achieve the highest C-Index, its performance is consistently close to the strongest baseline on each cohort. This result suggests that the gains in subtype separation do not come at the expense of predictive usefulness. Rather, AdaCSM improves survival stratification while maintaining strong rank-based prognostic performance. As a result, AdaCSM remains on the Pareto frontier of the trade-off between C-index and Logrank (Appendix Figure 7).

\subsection{Benefit of Sparse Expert Routing}
We next examine how expert count and routing sparsity affect performance. Here, we take the optimal hyperparameters from the dense AdaCSM model trained for each dataset (Appendix Tables 1-5).
\zw{\textbf{Figure~\ref{fig:sensitivity-logrank-plot}}} shows that subtype separation, measured by LogRank, generally improves when the model uses a moderate or large number of experts together with sparse or moderately sparse routing. The best configurations are dataset-dependent, but the strongest results are consistently obtained with 16 or 32 experts and relatively small top-$k$ values. For example, the best LogRank values are achieved with 32 experts, $k=4$ on the PBC dataset (311.51), 32 experts, $k$=4 on the FLCHAIN dataset (628.28). These findings indicate that increasing expert diversity is beneficial, but only when routing remains selective enough to preserve specialization.

Figure~\ref{fig:sensitivity-cindex-plot} presents the corresponding sensitivity analysis using the C-Index. Compared with LogRank, the variation across expert configurations is smaller, indicating that routing choices affect subtype separation more strongly than rank-based risk prediction. The best C-Index values are 0.6736 on SUPPORT, 0.7944 on PBC, 0.7574 on Framingham, and 0.7927 on FLCHAIN. This suggests that the predictive accuracy of AdaCSM is relatively sensitive to architectural choices, while its clustering quality is more robust to the degree of expert specialization.

Sparse routing also reduces computational cost because only a subset of experts is activated for each input. Table~\ref{tab:model_complexity} compares the computational complexity of AdaCSM with classical and deep survival baselines using the optimal hyperparameter configurations obtained for the SUPPORT dataset, which has the largest feature dimensionality among the evaluated cohorts. We report the number of active parameters and approximate floating-point operations (FLOPs) required per sample during inference, as well as total parameters for the maximum model capacity. For dense neural models such as DeepSurv, DSM, and DCSM, all parameters are activated for every input. In contrast, while AdaCSM contains a larger number of total parameters due to the presence of multiple expert networks, the effective computation during inference is controlled by the Top-$k$ sparsity routing mechanism, where only a small subset of experts are activated for each sample. This allows AdaCSM to have significantly fewer active parameters and lower FLOPs when sparse routing is used. For example, with $k=1$, AdaCSM activates only 4.97K effective parameters and requires 9.72K FLOPs per sample, which is lower than all dense neural survival models, while achieving a LogRank score that outperforms the best-performing baseline model (Figures~\ref{fig:logrank-plot} and \ref{fig:sensitivity-logrank-plot}). Increasing $k$ increases the computational cost but allows more experts to contribute to the representation. This conditional computation enables AdaCSM to maintain a higher representational capacity while keeping the effective inference cost comparable to or lower than conventional dense models.

\begin{table}[htbp]
\centering
\caption{Complexity comparison across survival models.}
\label{tab:model_complexity}
\small
\begin{tabular}{lccc}
\toprule
Model & Active Params & FLOPs/sample & Total Params \\
\midrule
CoxPH & 90 & 180 & 90 \\

AdaCSM & 18.47K & 36.72K & 18.47K \\
AdaCSM ($k=2$) & 9.47K & 18.72K & 18.47K \\
AdaCSM ($k=1$) & 4.97K & 9.72K & 18.47K \\

DeepCoxPH & 7.15K & 14.20K & 7.15K \\
DSM & 7.10K & 14.20K & 7.10K \\
DCSM & 7.10K & 14.20K & 7.10K \\
\bottomrule
\end{tabular}
\end{table}

Overall, sparse and moderately sparse routing ($k$ = 1, 2, 4) with a moderate number of experts (16-32) provides stable performance across datasets. The general trend indicates that balanced expert capacity and routing sparsity offer the most robust performance. We find that the behavior is broadly consistent with observations in prior MoE architectures, where sparse routing offers balanced efficiency and predictive stability.

\fz{\subsection{Ablation Studies}}

\begin{table}[ht]
\centering
\caption{Ablation study of the gating mechanism on the PBC dataset. We evaluate the impact of adaptive routing compared to non-informed weighting strategies on prognostic accuracy (C-Index) and subtype separation (LogRank).}
\label{tab:ablation_pbc}
\small
\begin{tabular}{lcc}
\toprule
\textbf{Configuration} & \textbf{C-Index ($\uparrow$)} & \textbf{LogRank ($\uparrow$)} \\
\midrule
Random Weights & $0.7366 \pm 0.0039$ & $75.60 \pm 56.83$ \\
Equal Weights (Uniform) & $0.7827 \pm 0.0195$ & $64.92 \pm 85.23$ \\
Single Expert ($n=1$) & $0.8131 \pm 0.0162$ & $254.65 \pm 39.16$ \\
\midrule
\textbf{AdaCSM (Full)}   & $\mathbf{0.8218 \pm 0.0084}$ & $\mathbf{311.80 \pm 19.30}$ \\
\bottomrule
\end{tabular}
\end{table}

\fz{
To evaluate the individual contributions and the adaptive gating mechanism of the MoE component in AdaCSM framework, we conducted ablation experiments on the PBC dataset (Table~\ref{tab:ablation_pbc}). We compare the AdaCSM framework against three constrained baselines. First, we ran AdaCSM with random weights to the experts, where expert routing is non-informed. Second, we assigned equal weights, where all experts contribute equally to the final representation. Lastly, we employed a single expert baseline, representing a standard deep survival model without mixture components. The results show that the adaptive gating mechanism is a critical driver of AdaCSM's performance and stability. Replacing the learned routing with random weights leads to a worse drop in prognostic accuracy than subtyping performance. The equal weights baseline results in highly unstable subtype separation, showing low LogRank but high standard deviation. In contrast, AdaCSM achieves a superior and stable LogRank and C-index. These findings confirm that the gating network in AdaCSM is the driver of its performance. The performance gap between the single expert baseline and AdaCSM further justifies the use of the MoE framework for handling the high variance and heterogeneity present in clinical survival cohorts.
}

\subsection{Expert Specialization Across Clinical Subgroups}
\begin{figure}[t]
\centering
\includegraphics[width=0.8\linewidth]{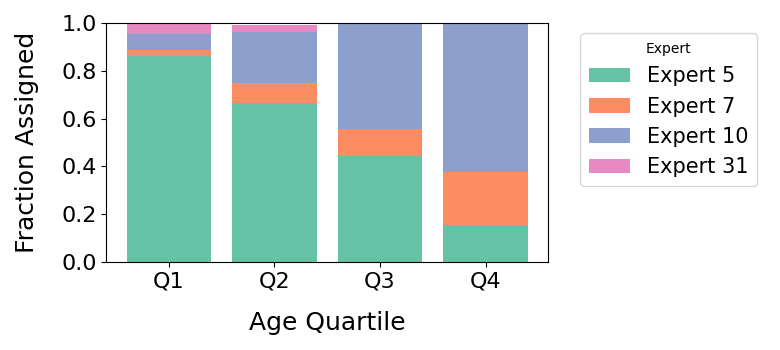}

\includegraphics[width=0.8\linewidth]{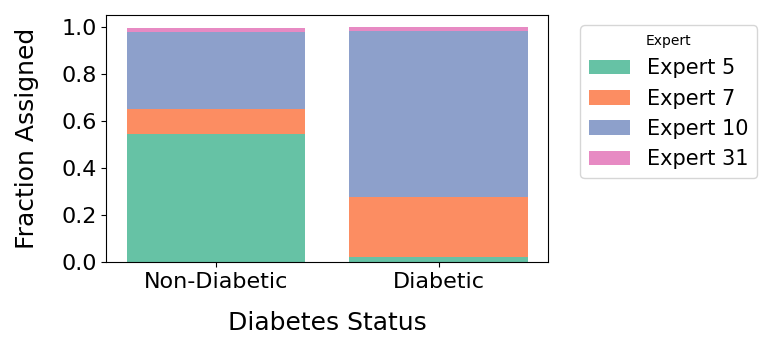}

\includegraphics[width=0.8\linewidth]{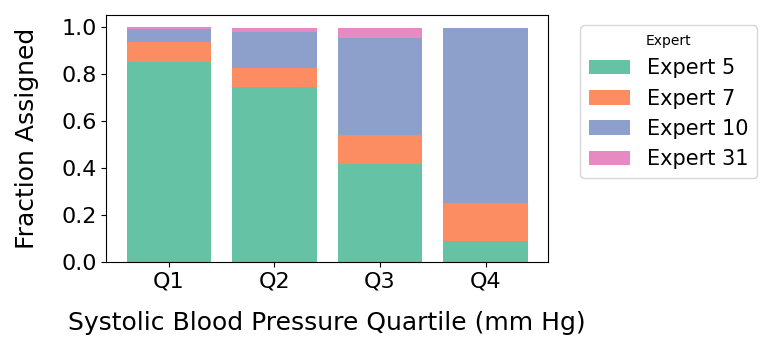}

    \caption{Expert specialization analysis on Framingham. Top: Expert assignments vary by age, diabetes, and blood pressure. Bottom: Clinical feature distributions for Expert 5 (lower-risk) and Expert 10 (higher-risk profile).}
    \label{fig:combined_expert_analysis}

\vspace{-20pt}
\end{figure}

\begin{figure}[htbp]
    \centering
    \begin{minipage}{0.48\textwidth}
        \centering
        \includegraphics[width=0.7\linewidth]{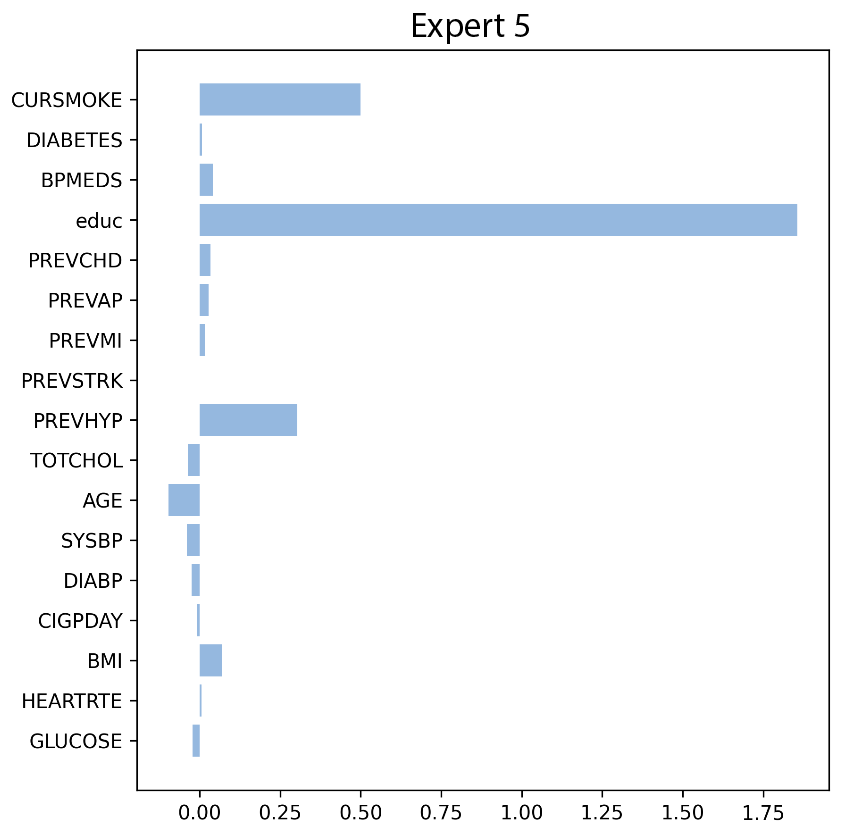}
    \end{minipage}
    \hfill
    \begin{minipage}{0.48\textwidth}
        \centering
        \includegraphics[width=0.7\linewidth]{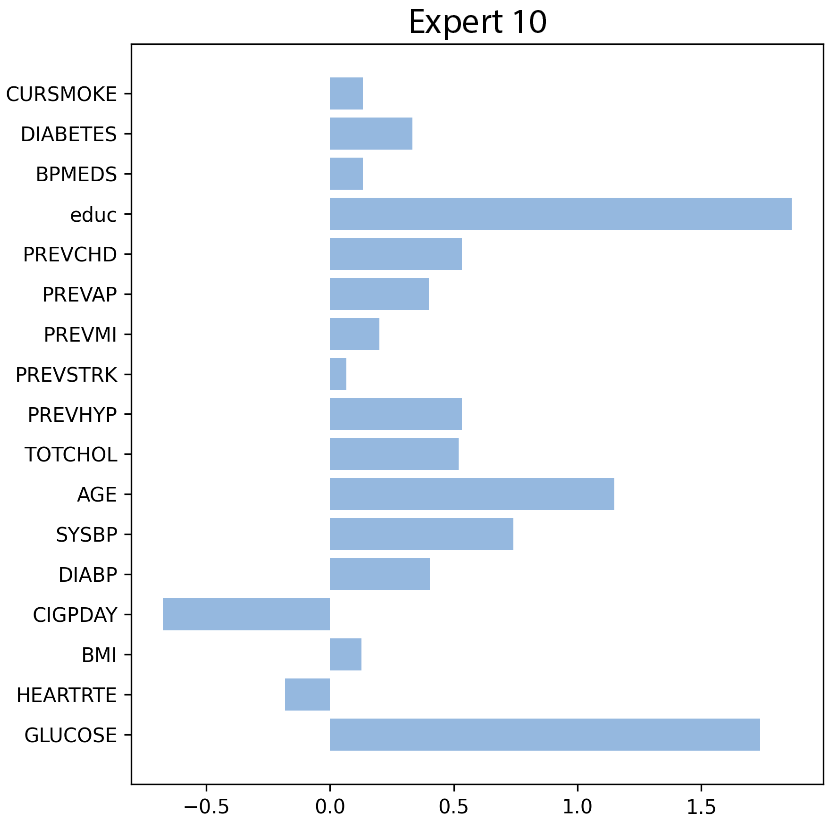}
    \end{minipage}
    \caption{Clinical feature distributions for Expert 5 (high risk) and Expert 10 (low risk). }
    \label{fig:expert_specialization}
\vspace{-15pt}
\end{figure}

To better understand how sparse AdaCSM leverages the mixture-of-experts architecture, we examine whether the routing mechanism learns clinically meaningful expert specialization. In particular, we analyze how the gating network assigns patients to experts across clinically relevant subgroups.  We hypothesize that a meaningful routing mechanism would show experts to specialize in different patient populations rather than assigning samples uniformly. Here, we use the Framingham cohort for the study of cardiovascular disease using mortality as an outcome in this case study.

Figure~\ref{fig:combined_expert_analysis} illustrates the expert assignment patterns learned by sparse AdaCSM. The routing distribution varies systematically across age groups, diabetes status, and systolic blood pressure. Younger, low blood pressure, and non-diabetic subjects are routed primarily to Expert 5, whereas older, high blood pressure, and diabetic subjects are assigned more frequently to Expert 10. The distinct specializations by the experts follow clinically relevant and interpretable subgroups for risks of mortality. These findings suggest that AdaCSM does not use experts uniformly but instead allocates different clinically interpretable subgroups to experts in a structured way. This behavior supports the interpretation that the MoE encoder captures subgroup-specific feature patterns that may contribute to improved survival stratification. In a clinical setting, this model behavior is desirable since patient populations are often heterogeneous and risk factors may interact differently across demographic and comorbidity groups. 

While the routing patterns show that expert specialization occurs,  Figure~\ref{fig:expert_specialization} further reveals the overall combination of patient conditions or the clinical logic that drives the subtype assignments. By inspecting the feature distributions, we observe that Expert 5 focuses on a protective profile, characterized by lower glucose levels and blood pressure. On the other hand, Expert 10 characterizes a high-risk profile where metabolic, cardiovascular markers, and pre-existing conditions are elevated. This deep dive into the expert profiles confirms that the gating mechanism by AdaCSM clusters patients in a clinically meaningful and interpretable way, identifying specific physiological states associated with different survival trajectories. In this case study, since the outcome of our data is mortality, factors contributing to the high risk preference of Expert 5 might be apparent when compared to the low-risk preference of Expert 10. 

The dynamic routing of patients to specialized experts is consistently observed across all evaluated datasets. We provide additional expert assignment profiles for the PBC, SUPPORT, and FLCHAIN cohorts in Appendix A.7 (Appendix Figures 4, 5, and 6). These confirm that the adaptive gating network in AdaCSM captures known drivers for mortality in each cohort, such as liver enzymes in the liver disease cohort from the PBC dataset (Appendix Figure 4). However, as clinical outcomes become more complex and as subtypes of interest increase, for example, when predicting the outcome or stages of a disease, the granular interpretability of expert association to subtype assignments that AdaCSM offers will be essential for understanding the underlying combinations of risk factors. 

To further validate that these learned expert specializations are grounded in robust feature importance, we compare the intrinsic expert profiles against post-hoc SHAP~\cite{lundberg2017shap} attributions. We observe a high correlation between our learned expert profiles and SHAP importance, hence confirming that the gating mechanism consistently routes patients based on clinically relevant features (Appendix Figure 7, Appendix A.8).

\section{Discussion}
Overall, the results suggest that AdaCSM is particularly effective at identifying survival subtypes. Across all four clinical cohorts, the model achieves stronger LogRank performance than the competing baselines, which means that the discovered groups are more clearly separated in terms of survival trajectory. At the same time, AdaCSM remains competitive in C-Index, so the improvement in subtype stratification does not come with a major loss in predictive performance. This pattern is important because our main goal is not only to rank patients by risk, but also to recover clinically meaningful subgroups with distinct prognoses. 

The substantial improvement comes from the MoE encoder. In the original DCSM formulation, all patients are passed through a single shared encoder, which may not be flexible enough to represent heterogeneous clinical populations. In AdaCSM, the gating network routes different patients to different experts, allowing the model to learn more specialized representations for different survival patterns. This is consistent with the design of our model, where the MoE-enhanced representation is used to estimate subtype weights and combine subtype-specific Weibull survival functions into the final prediction. \fz{Future work could further extend the intrinsic interpretability of our framework. The modular experts provide a structured interface for integration with large language models (LLMs) to provide natural language explanations of the personalized patient routing and specializations for clinical interpretability. This allows for evidence-based clinical reports that could be generated when explaining a patient's subtype assignment.}

The sensitivity analysis also helps explain when the model works best. In general, increasing the number of experts improves performance up to a point, especially when routing is sparse. However, using too many experts together with dense routing can slightly hurt performance. We see a fairly consistent trend that sparse or moderately sparse routing, especially with \(k=1,2,\) or \(4\), gives the most stable results across datasets. This suggests that the benefit of AdaCSM is not simply larger model capacity, but the combination of enough expert diversity with enough routing sparsity to preserve specialization. 

There are still several limitations. The current model fixes the number of subtypes to two, which aligns with clinical interpretability but may miss more complex disease structures. Extending the framework to automatically determine the number of clusters and evaluate larger numbers of subtypes is an important direction for future work. In addition, the Weibull survival experts impose a parametric assumption, and the current experiments do not yet fully explain the clinical meaning of individual experts. Future work will explore multimodal extensions, temporal routing, and transfer of expert modules across related disease settings.


\section{Conclusion}
In this work, we propose AdaCSM, a deep clustering survival model based on a mixture-of-experts framework to better capture clinical heterogeneity in survival prediction. Our approach uses a learned gating mechanism to route patients to specialized neural experts. Hence, allowing distinct disease progression patterns to be modeled within a unified framework. We evaluate and demonstrate the effectiveness of the proposed model on real-world clinical survival datasets. The experimental results show consistent improvements in survival and subtype clustering performance compared to existing deep survival models. We also evaluate the interpretability of AdaCSM with case studies. We further analyze the gating behavior to examine how expert specialization relates to clinical subgroups. 
An additional advantage of this design is that the computational cost of AdaCSM can be adjusted through the routing sparsity parameter $k$. In practice, this provides flexibility for deployment across clinical environments with different computational constraints. For example, smaller values of $k$ with reduced computational cost make the model suitable for resource-constrained settings, while larger values allow more experts to participate when higher computational budgets are available.
Future directions include extending the model to support multimodal data fusion. Furthermore, we aim to investigate temporal routing mechanisms, such as with longitudinal recurrent layers into the gating network, to allow the model to adapt its subtype assignments as clinical biomarkers shift over time. Finally, we plan to explore the transferability of our modular experts across other neurodegenerative disorders to determine if the identified survival kernels capture generalized biological mechanisms of neurodegeneration.

\begin{acks}

This work was supported in part by the NIH grants U01 AG068057, U01 AG066833, and U19 AG074879; and NSF grant 2500343. 
\end{acks}



\bibliographystyle{ACM-Reference-Format}
\bibliography{references}

@article{knaus1995support,
  title={The SUPPORT prognostic model: Objective estimates of survival for seriously ill hospitalized adults},
  author={Knaus, William A and Harrell, Frank E and Lynn, Joanne and Goldman, Lee and Phillips, Russell S and Connors, Alfred F and Dawson, Neal V and Fulkerson, William J and Califf, Robert M and Desbiens, Norman and others},
  journal={Annals of internal medicine},
  volume={122},
  number={3},
  pages={191--203},
  year={1995},
  publisher={American College of Physicians}
}

@book{fleming2013PBC,
  title={Counting processes and survival analysis},
  author={Fleming, Thomas R and Harrington, David P},
  year={2013},
  publisher={John Wiley \& Sons}
}

@article{dawber1951framingham,
  title={Epidemiological approaches to heart disease: the Framingham Study},
  author={Dawber, Thomas R and Meadors, Gilcin F and Moore Jr, Felix E},
  journal={American Journal of Public Health and the Nations Health},
  volume={41},
  number={3},
  pages={279--286},
  year={1951},
  publisher={American Public Health Association}
}

@inproceedings{dispenzieri2012FLCHAIN,
  title={Use of nonclonal serum immunoglobulin free light chains to predict overall survival in the general population},
  author={Dispenzieri, Angela and Katzmann, Jerry A and Kyle, Robert A and Larson, Dirk R and Therneau, Terry M and Colby, Colin L and Clark, Raynell J and Mead, Graham P and Kumar, Shaji and Melton III, L Joseph and others},
  booktitle={Mayo Clinic Proceedings},
  volume={87},
  number={6},
  pages={517--523},
  year={2012},
  organization={Elsevier}
}

@article{lundberg2017shap,
  title={A unified approach to interpreting model predictions},
  author={Lundberg, Scott M and Lee, Su-In},
  journal={Advances in neural information processing systems},
  volume={30},
  year={2017}
}

@article{wei1992accelerated,
  title={The accelerated failure time model: a useful alternative to the Cox regression model in survival analysis},
  author={Wei, Lee-Jen},
  journal={Statistics in medicine},
  volume={11},
  number={14-15},
  pages={1871--1879},
  year={1992},
  publisher={Wiley Online Library}
}

@article{buckley1979linear,
  title={Linear regression with censored data},
  author={Buckley, Jonathan and James, Ian},
  journal={Biometrika},
  volume={66},
  number={3},
  pages={429--436},
  year={1979},
  publisher={Oxford University Press}
}

@article{ishwaran2008random,
  title={Random survival forests},
  author={Ishwaran, Hemant and Kogalur, Udaya B and Blackstone, Eugene H and Lauer, Michael S},
  year={2008}
}

@article{ishwaran2014random,
  title={Random survival forests for competing risks},
  author={Ishwaran, Hemant and Gerds, Thomas A and Kogalur, Udaya B and Moore, Richard D and Gange, Stephen J and Lau, Bryan M},
  journal={Biostatistics},
  volume={15},
  number={4},
  pages={757--773},
  year={2014},
  publisher={Oxford University Press}
}

@article{bland1998survival,
  title={Survival probabilities (the Kaplan-Meier method)},
  author={Bland, J Martin and Altman, Douglas G},
  journal={Bmj},
  volume={317},
  number={7172},
  pages={1572--1580},
  year={1998},
  publisher={British Medical Journal Publishing Group}
}

@inproceedings{alaa2017deep,
  title={Deep multi-task gaussian processes for survival analysis with competing risks},
  author={Alaa, Ahmed M and van der Schaar, Mihaela},
  booktitle={Proceedings of the 31st International Conference on Neural Information Processing Systems},
  pages={2326--2334},
  year={2017}
}

@article{hartigan1979algorithm,
  title={Algorithm AS 136: A k-means clustering algorithm},
  author={Hartigan, John A and Wong, Manchek A},
  journal={Journal of the royal statistical society. series c (applied statistics)},
  volume={28},
  number={1},
  pages={100--108},
  year={1979},
  publisher={JSTOR}
}

@article{cox1972regression,
  title={Regression models and life-tables},
  author={Cox, David R},
  journal={Journal of the royal statistical society: Series B (methodological)},
  volume={34},
  number={2},
  pages={187--202},
  year={1972},
  publisher={Wiley Online Library}
}

@article{katzman2018deepsurv,
  title={DeepSurv: personalized treatment recommender system using a Cox proportional hazards deep neural network},
  author={Katzman, Jared L and Shaham, Uri and Cloninger, Alexander and Bates, Jonathan and Jiang, Tingting and Kluger, Yuval},
  journal={BMC medical research methodology},
  volume={18},
  number={1},
  pages={24},
  year={2018},
  publisher={Springer}
}

@article{nagpal2021deep,
  title={Deep survival machines: Fully parametric survival regression and representation learning for censored data with competing risks},
  author={Nagpal, Chirag and Li, Xinyu and Dubrawski, Artur},
  journal={IEEE Journal of Biomedical and Health Informatics},
  volume={25},
  number={8},
  pages={3163--3175},
  year={2021},
  publisher={IEEE}
}

@article{hou2024DCSM,
  title={Interpretable deep clustering survival machines for Alzheimer’s disease subtype discovery},
  author={Hou, Bojian and Wen, Zixuan and Bao, Jingxuan and Zhang, Richard and Tong, Boning and Yang, Shu and Wen, Junhao and Cui, Yuhan and Moore, Jason H and Saykin, Andrew J and others},
  journal={Medical image analysis},
  volume={97},
  pages={103231},
  year={2024},
  publisher={Elsevier}
}

@inproceedings{hou2023deep,
  title={Deep clustering survival machines with interpretable expert distributions},
  author={Hou, Bojian and Li, Hongming and others},
  booktitle={2023 IEEE 20th International Symposium on Biomedical Imaging (ISBI)},
  pages={1--4},
  year={2023},
  organization={IEEE}
}

@inproceedings{wen2025multi,
  title={Multi-Modal Deep Clustering Survival Machines for Alzheimer's Disease Subtype Discovery},
  author={Wen, Zixuan and Hou, Bojian and others},
  booktitle={Proc. of the IEEE/CVF Int. Conf. on Computer Vision},
  pages={2264--2272},
  year={2025}
}

@inproceedings{chapfuwa2020survival,
  title={Survival cluster analysis},
  author={Chapfuwa, Paidamoyo and Li, Chunyuan and Mehta, Nikhil and Carin, Lawrence and Henao, Ricardo},
  booktitle={Proceedings of the ACM Conference on Health, Inference, and Learning},
  pages={60--68},
  year={2020}
}

@article{manduchi2021deep,
  title={A deep variational approach to clustering survival data},
  author={Manduchi, Laura and Marcinkevi{\v{c}}s, Ri{\v{c}}ards and Massi, Michela C and Weikert, Thomas and Sauter, Alexander and Gotta, Verena and M{\"u}ller, Timothy and Vasella, Flavio and Neidert, Marian C and Pfister, Marc and others},
  journal={arXiv preprint arXiv:2106.05763},
  year={2021}
}

@inproceedings{jeanselme2022neural,
  title={Neural Survival Clustering: Non-parametric mixture of neural networks for survival clustering},
  author={Jeanselme, Vincent and Tom, Brian and Barrett, Jessica},
  booktitle={Conference on Health, Inference, and Learning},
  pages={92--102},
  year={2022},
  organization={PMLR}
}

@article{qiu2025deep,
  title={Deep representation learning for clustering longitudinal survival data from electronic health records},
  author={Qiu, Jiajun and Hu, Yao andothers},
  journal={Nature Communications},
  volume={16},
  number={1},
  pages={2534},
  year={2025},
  publisher={Nature Publishing Group UK London}
}

@inproceedings{noshin2025integrating,
  title={Integrating social determinants of health in a multi-modal deep clustering survival model for injury-risk in alzheimer’s and related dementia patients},
  author={Noshin, Kazi and Boland, Mary Regina and Hou, Bojian and He, Weiqing and Lu, Victoria and Shen, Li and Zhang, Aidong},
  booktitle={AAAI},
  year={2025}
}

@inproceedings{akiba2019optuna,
  title={Optuna: A next-generation hyperparameter optimization framework},
  author={Akiba, Takuya and Sano, Shotaro and Yanase, Toshihiko and Ohta, Takeru and Koyama, Masanori},
  booktitle={Proceedings of the 25th ACM SIGKDD international conference on knowledge discovery \& data mining},
  pages={2623--2631},
  year={2019}
}

\appendix
\gdef\thefigure{\thesection.\arabic{figure}}
\setcounter{figure}{0}

\gdef\thetable{\thesection.\arabic{table}}
\setcounter{table}{0}

\gdef\theequation{\thesection.\arabic{equation}}
\setcounter{equation}{0}

\appendix

\section{Appendix}

\subsection{Data Preprocessing}
\label{sec:data_preprocessing}
This section details the cohort-specific preprocessing steps used to arrive at the final statistics presented in Table 1.

To ensure training stability across datasets measured in different units (e.g., days vs. years), we further apply a global max-scaling transformation to the survival times. For each dataset, every observed time $t$ is divided by the maximum time value present in the training split:
\begin{equation}
    t_{scaled} = \frac{t}{\max(t_{train})}
\end{equation}

\subsubsection{\textbf{SUPPORT}}
The Study to Understand Prognoses and Preferences for Outcomes and Risks of Treatments (SUPPORT) dataset consists of 9,105 patients. As each patient appears only once in the cohort, no row-level exclusions were required. Preprocessing followed these steps:
\begin{itemize}
    \item \textbf{Feature Selection and Leakage Prevention:} To ensure the model predicts risk from baseline patient factors rather than existing clinical scores or outcomes, we removed hospital death indicators (\texttt{hospdead}), existing composite severity-of-illness scores (\texttt{aps, sps}), and subjective physician survival estimates (\texttt{surv2m, surv6m}). 
    \item \textbf{Skewness Correction:} To handle strong right-skewness and stabilize training, we applied a log-transformation to cost-related variables and the oxygenation index (\texttt{PaO$_2$/FiO$_2$}).
    \item \textbf{Imputation Strategy:} Missing continuous predictors were imputed using the median of the training data. For missing categorical variables, we utilized the mode of the training distribution.
    \item \textbf{Encoding and Scaling:} Categorical variables—including sex, primary disease group, disease classification, income bracket, race, presence of malignancy, and functional status (SF-20)—were one-hot encoded. All continuous features were subsequently standardized to a zero mean and unit variance.
    \item \textbf{Target Labeling:} The binary \texttt{death} column was used as the ground-truth event indicator ($0 = \text{alive}, 1 = \text{dead}$).
\end{itemize}

\subsubsection{\textbf{PBC}}
The Primary Biliary Cholangitis (PBC) dataset contains clinical measurements with 1,945 total observations. To align this with the baseline prognostic task, we performed the following preprocessing steps:
\begin{itemize}
    \item \textbf{Target Labeling:} We utilized the \texttt{status2} variable as the ground-truth event indicator ($0 = \text{alive}, 1 = \text{dead}$). The original \texttt{status} column, which includes liver transplantation as a separate state, was excluded to focus specifically on all-cause mortality.
    \item \textbf{Feature Encoding:} Categorical clinical markers, including \textit{drug, sex, ascites, hepatomegaly, spiders, edema,} and \textit{histologic stage}, were transformed using one-hot encoding. Continuous features were standardized to zero mean and unit variance using the training set statistics.
    \item \textbf{Feature Dimensionality:} After expanding categorical variables, the final processed dataset consists of $d=25$ input features.
    \item \textbf{Missing Data:} At baseline, we observed 9.0\% missingness for serum cholesterol and 1.3\% for platelets. These missing continuous predictors were imputed using the column mean of the train data prior to modeling.
\end{itemize}

\subsubsection{\textbf{Framingham}}
The Framingham Heart Study dataset contains 11,627 exam records. To create a baseline prognostic model, we processed the data as follows:
\begin{itemize}
    \item \textbf{Feature Selection and Leakage Prevention:} To prevent data leakage, we excluded all follow-up clinical event indicators and their corresponding timestamps (e.g., \textit{Angina, Stroke, Myocardial Infarction, Hypertension, and CVD} flags). The model predictors were restricted to baseline demographics and physiological measurements.
    \item \textbf{Handling Lipids:} Although \textit{Total Cholesterol} (\texttt{TOTCHOL}) was retained, the specific sub-fractions \textit{HDL} and \textit{LDL} were excluded from the final feature set due to 100\% missingness in the baseline exam rows of the provided source.
    \item \textbf{Imputation Strategy:} Missing continuous values for variables such as \textit{Glucose} (8.95\% missingness), \textit{BPMEDS}, and \textit{BMI} were imputed using the column median. For categorical variables, missing values were handled during one-hot encoding by assigning them to an all-zero vector across the dummy columns for that field.
    \item \textbf{Encoding and Scaling:} Ten categorical fields—including \textit{sex, smoking status, diabetes, education level, and medical histories (e.g., previous stroke or heart disease)}—were one-hot encoded. All 10 numeric features were standardized to a zero mean and unit variance.
    \item \textbf{Target Labeling:} The survival outcome was defined as the time from the baseline exam to either death or the end of the study period (\texttt{TIMEDTH} - \texttt{TIME}). The \texttt{DEATH} column served as the binary event indicator ($0 = \text{alive}, 1 = \text{dead}$).
\end{itemize}

\subsubsection{\textbf{FLCHAIN}}
The Medical College of Wisconsin's Free Light Chain (FLCHAIN) dataset consists of 6,524 subjects. As the source data provides a single baseline record per individual, no row-level filtering was required. The following preprocessing steps were applied:
\begin{itemize}
    \item \textbf{Feature Representation:} We utilized eight predictors, including demographics (\textit{age, sex}), clinical markers (\textit{kappa, lambda, creatinine}), and research-specific variables (\textit{sample year, FLC group index, MGUS status}). Notably, \textit{sex} and \textit{FLC group} were treated as numerical/ordinal inputs rather than undergoing one-hot encoding.
    \item \textbf{Missing Data and Scaling:} No missing values were detected in the predictor columns for this cohort. Consistent with the other datasets, a \textit{StandardScaler} was applied to all eight features to normalize the input distribution based on train data.
    \item \textbf{Time-to-Event Processing:} To prevent zero-length follow-up intervals during model optimization, a +1 day shift was applied to the follow-up duration (\texttt{futime}).
    \item \textbf{Target Labeling:} The survival endpoint was defined as all-cause mortality, using the \texttt{death} column as the binary event indicator ($0 = \text{alive}, 1 = \text{dead}$).
\end{itemize}

\subsection{Optuna Hyperparameter Tuning}\label{sec:optuna}

Hyperparameters for all models were selected using Optuna. For each method and dataset, we defined a model-specific search space over key architectural and optimization parameters, such as learning rate, hidden layer size, batch size, dropout, and method-specific settings. Optuna was then used to search for the configuration that maximized the validation C-Index. The best hyperparameter settings for AdaCSM, Cox PH, DeepSurv, DSM, and DCSM are reported in Tables 1-5, respectively.

\begin{table*}[h]
\centering
\caption{Best hyperparameters selected for AdaCSM on each clinical survival dataset using Optuna tuning.}
\label{tab:dense_adacsm_hyperparams}
\small
\begin{tabular}{lcccc}
\toprule
Hyperparameter & SUPPORT & PBC & Framingham & FLCHAIN \\
\midrule
Learning rate & 1.50e-4 & 4.68e-4 & 6.21e-3 & 3.38e-3 \\
Discount & 0.6655 & 0.8316 & 0.5659 & 0.9413 \\
Hidden layers & [50] & [100] & [100] & [50] \\
Number of experts & 4 & 32 & 32 & 32 \\
Batch size & 16 & 16 & 100 & 100 \\
Dropout & 0.0294 & 0.0477 & 0.0544 & 0.1447 \\
Gate dropout & 0.0609 & 0.2708 & 0.0982 & 0.0093 \\
Gate temperature & 3.0002 & 0.1191 & 0.8553 & 0.8817 \\
Load balance $\lambda$ & 0.0294 & 0.0952 & 0.0725 & 0.0836 \\
\bottomrule
\end{tabular}
\end{table*}
\begin{table*}[h]
\centering
\caption{Best hyperparameters selected for Cox Proportional Hazards (Cox PH) on each clinical survival dataset using Optuna tuning.}
\begin{tabular}{lcccc}
\toprule
Hyperparameter & SUPPORT & PBC & Framingham & FLCHAIN \\
\midrule
Penalizer & 0.0096 & 0.0894 & 0.0019 & 0.0097 \\
L1 ratio  & 0.3594 & 0.6936 & 0.4612 & 0.3861 \\
\bottomrule
\end{tabular}
\label{tab:coxph_hyperparams}
\end{table*}

\begin{table*}[h]
\centering
\caption{Best hyperparameters for DeepSurv on clinical survival datasets.}
\label{tab:deepsurv_hyperparams}
\small
\begin{tabular}{lcccc}
\toprule
Hyperparameter & SUPPORT & PBC & FRAMINGHAM & FLCHAIN \\
\midrule

Learning Rate & 1.31e-4 & 2.14e-4 & 7.06e-4 & 5.88e-3 \\
Hidden Layers & [100] & [50,50] & [100] & [50] \\
Batch Size & 100 & 100 & 16 & 100 \\
\bottomrule
\end{tabular}
\end{table*}

\begin{table*}[h]
\centering
\caption{Best hyperparameters for Deep Survival Machines (DSM) on clinical survival datasets.}
\label{tab:dsm_hyperparams_updated}
\small
\begin{tabular}{lcccc}
\toprule
Hyperparameter & SUPPORT & PBC & Framingham & FLCHAIN \\
\midrule
Learning Rate & 1.67e-4 & 9.04e-3 & 1.54e-3 & 3.79e-5 \\
Hidden Layers & [50, 50] & [50, 50] & [50, 50] & [50] \\
Batch Size & 32 & 32 & 16 & 128 \\
Discount & 0.8990 & 0.4433 & 0.4655 & 0.6438 \\
Distribution & Weibull & Weibull & Weibull & Weibull \\
\bottomrule
\end{tabular}
\end{table*}

\begin{table*}[h]
\centering
\caption{Best hyperparameters for Deep Clustering Survival Machines (DCSM) on clinical survival datasets.}
\label{tab:dcsm_hyperparams}
\small
\begin{tabular}{lcccc}
\toprule
Hyperparameter & SUPPORT & PBC & Framingham & FLCHAIN \\
\midrule
Learning rate & 1.33e-4 & 1.60e-4 & 4.01e-4 & 8.07e-3 \\
Discount & 0.4092 & 0.7662 & 0.6950 & 0.7760 \\
Hidden layers & [50] & [50] & [50, 50] & [100] \\
Batch size & 100 & 100 & 100 & 100 \\
\bottomrule
\end{tabular}
\end{table*}

\subsection{Subtype Clustering}
\zw{Table~\ref{tab:logrank_clinical} shows the LogRank results of different models. AdaCSM achieves the highest LogRank statistic on all four clinical cohorts. In particular, the improvement over DCSM is consistent across all datasets, suggesting that adaptive expert specialization enhances subtype discovery beyond the original clustering survival formulation.} 
\begin{table*}[!t]
\centering
\caption{LogRank statistic ($\chi^2$ $\uparrow$, mean $\pm$ std) measuring survival separation between discovered subtypes on clinical survival cohorts. For each dataset, the best performance is in bold, while the second-best is underlined.}
\label{tab:logrank_clinical}
\small
\begin{tabular}{lcccc}
\toprule
\textbf{Model} & \textbf{SUPPORT} & \textbf{PBC} & \textbf{FRAMINGHAM} & \textbf{FLCHAIN} \\
\midrule
Cox PH & 507.38 $\pm$ 52.86 & 192.17 $\pm$ 17.61 & 576.42 $\pm$ 21.97 & 401.75 $\pm$ 30.51 \\
DeepSurv & 584.00 $\pm$ 54.16 & 85.15 $\pm$ 115.00 & 590.78 $\pm$ 22.22 & 331.51 $\pm$ 187.13 \\
DSM & 879.19 $\pm$ 117.15 & 229.78 $\pm$ 49.58 & 613.96 $\pm$ 24.76 & 457.62 $\pm$ 116.75 \\
DCSM & 757.65 $\pm$ 46.64 & 292.54 $\pm$ 25.96 & 568.74 $\pm$ 44.43 & 563.08 $\pm$ 95.05 \\
AdaCSM & \textbf{1047.17 $\pm$ 114.04} & \underline{311.80 $\pm$ 19.30} & \textbf{667.83 $\pm$ 44.24} & \textbf{586.56 $\pm$ 38.39} \\
AdaCSM (Sparse top-2) & \underline{1028.72 $\pm$ 121.93} & \textbf{316.42 $\pm$ 38.88} & \underline{638.72 $\pm$ 44.60} & \underline{569.81 $\pm$ 28.61} \\
\bottomrule
\end{tabular}
\end{table*}

\subsection{Prognostic Accuracy}

\fz{
Table~\ref{tab:cindex_clinical} presents the global C-index performance across all four cohorts. These results demonstrate that AdaCSM maintains discriminative accuracy comparable to state-of-the-art dense models like DeepSurv and DCSM. This confirms that the transition to a sparse mixture-of-experts architecture provides significant gains in subtype separation (Table~\ref{tab:logrank_clinical}) and interpretability without compromising the model's fundamental ability to rank individual patient risk.
}

\begin{table*}[h] 
\centering
\caption{Global C-Index ($\uparrow$ mean $\pm$ std) comparison on clinical survival cohorts. For each dataset, the best performance is in bold, while the second-best is underlined.}
\label{tab:cindex_clinical}
\small
\begin{tabular}{lcccc}
\toprule
\textbf{Model} & \textbf{SUPPORT} & \textbf{PBC} & \textbf{FRAMINGHAM} & \textbf{FLCHAIN} \\ 
\midrule
Cox PH 
& \textbf{\boldmath $0.8414 \pm 0.0037$} 
& $0.8106 \pm 0.0152$ 
& \textbf{\boldmath $0.7420 \pm 0.0093$} 
& $0.7812 \pm 0.0165$ \\ 

DeepSurv 
& $0.8276 \pm 0.0028$ 
& \underline{$0.8182 \pm 0.0151$} 
& \underline{$0.7415 \pm 0.0083$} 
& \textbf{\boldmath $0.7891 \pm 0.0054$} \\ 

DSM 
& $0.8239 \pm 0.0029$ 
& $0.8090 \pm 0.0232$ 
& $0.7390 \pm 0.0080$ 
& \underline{$0.7881 \pm 0.0038$} \\ 

DCSM 
& $0.8151 \pm 0.0033$ 
& $0.8153 \pm 0.0138$ 
& $0.7293 \pm 0.0064$ 
& $0.7780 \pm 0.0079$ \\ 

AdaCSM (Sparse top-2)
& \underline{$0.8289 \pm 0.0052$} 
& \textbf{\boldmath $0.8218 \pm 0.0099$} 
& $0.7345 \pm 0.0036$ 
& $0.7845 \pm 0.0099$ \\ 

AdaCSM 
& $0.8284 \pm 0.0060$ 
& $0.8181 \pm 0.0142$ 
& $0.7300 \pm 0.0108$ 
& $0.7806 \pm 0.0088$ \\ 

\bottomrule
\end{tabular}
\end{table*}

\subsection{Time-dependent metrics}

\begin{figure*}
  \includegraphics[width=1\textwidth]{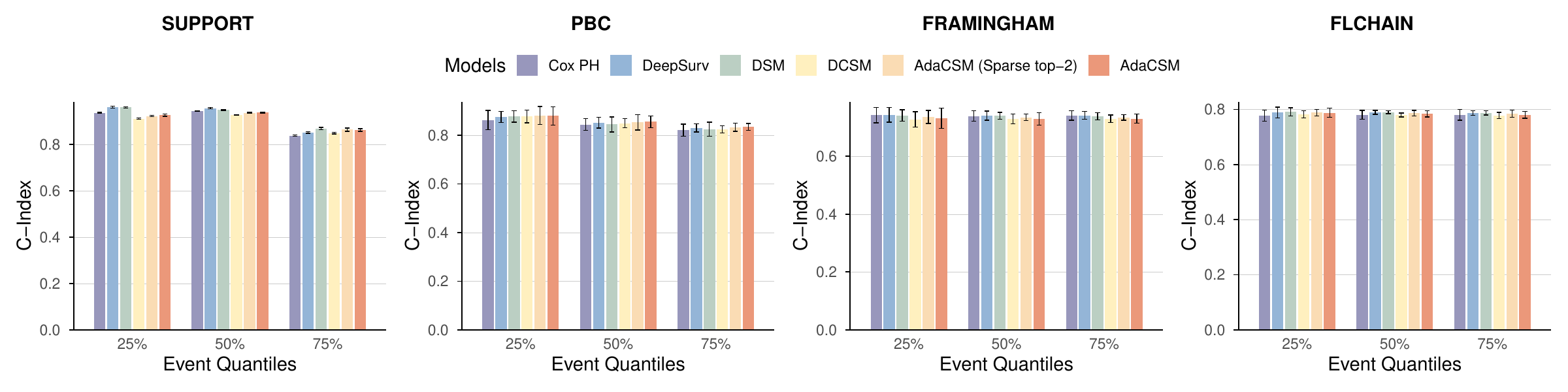}
  \caption{\zw{Bar plots of C-Index at the 25$\%$, 50$\%$, and 75$\%$ event quantiles for six survival models on the SUPPORT, PBC, FRAMINGHAM, and FLCHAIN datasets. Error bars represent 95$\%$ confidence intervals.}}
  \label{fig:cindex-quantiles-plot}
\end{figure*}

\begin{figure*}
  \includegraphics[width=1\textwidth]{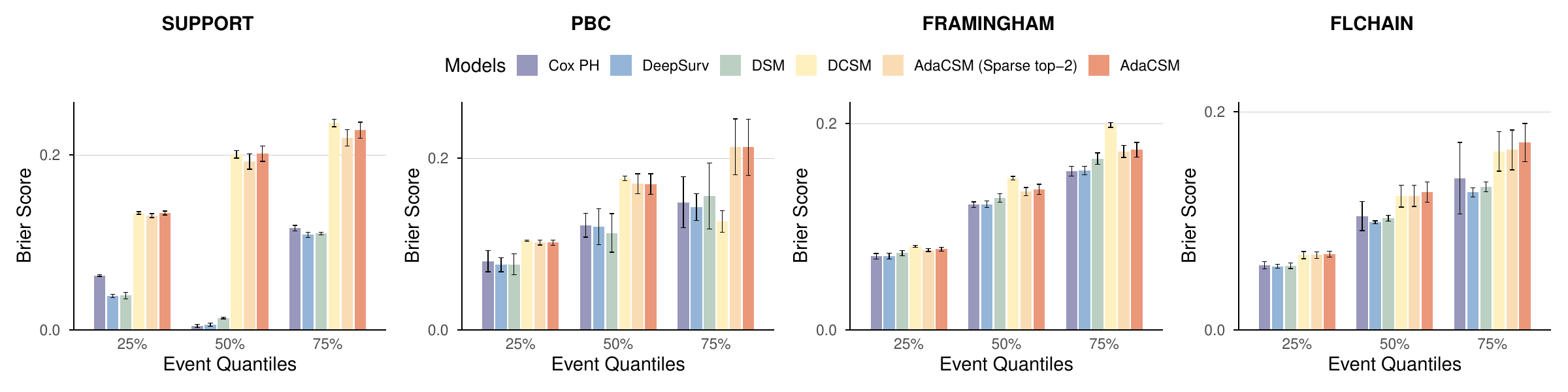}
  \caption{\zw{Bar plots of Brier Score at the 25$\%$, 50$\%$, and 75$\%$ event quantiles for six survival models on the SUPPORT, PBC, FRAMINGHAM, and FLCHAIN datasets. Error bars represent 95$\%$ confidence intervals.}}
  \label{fig:brier-quantiles-plot}
\end{figure*}

To evaluate model reliability over the clinical follow-up period, we report time-dependent C-index (Table~\ref{tab:cindex_quantiles_all}) and Brier Scores (Table~\ref{tab:brier_quantiles}) at the $25^{th}$, $50^{th}$, and $75^{th}$ event quantiles. AdaCSM exhibits stable longitudinal performance, particularly in the later stages of follow-up. The corresponding bar plots in Figures \ref{fig:cindex-quantiles-plot} and \ref{fig:brier-quantiles-plot} provide a visual comparison of these metrics, highlighting that our sparse routing mechanism maintains predictive consistency across varying survival horizons.

\begin{table*}[t]
\centering
\caption{C-Index ($\uparrow$ mean $\pm$ std) across different time quantiles ($25^{th}$, $50^{th}$, and $75^{th}$ percentiles) on clinical survival cohorts.}
\label{tab:cindex_quantiles_all}
\small
\begin{tabular}{lcccc}
\toprule
\textbf{Model / Quantile} & \textbf{SUPPORT} & \textbf{PBC} & \textbf{FRAMINGHAM} & \textbf{FLCHAIN} \\ \midrule
\textit{25th Percentile} & & & & \\
Cox PH & 0.9368 $\pm$ 0.0013 & 0.8630 $\pm$ 0.0320 & \underline{0.7422 $\pm$ 0.0212} & 0.7770 $\pm$ 0.0163 \\
DeepSurv & \textbf{\boldmath $0.9609 \pm 0.0028$} & 0.8754 $\pm$ 0.0179 & \textbf{\boldmath $0.7435 \pm 0.0205$} & \underline{0.7887 $\pm$ 0.0155} \\
DSM & \underline{0.9596 $\pm$ 0.0027} & \underline{0.8775 $\pm$ 0.0190} & 0.7412 $\pm$ 0.0161 & \textbf{\boldmath $0.7913 \pm 0.0119$} \\
DCSM & 0.9116 $\pm$ 0.0022 & 0.8777 $\pm$ 0.0211 & 0.7280 $\pm$ 0.0213 & 0.7818 $\pm$ 0.0104 \\
AdaCSM & 0.9267 $\pm$ 0.0044 & 0.8800 $\pm$ 0.0300 & 0.7320 $\pm$ 0.0285 & 0.7881 $\pm$ 0.0130 \\
AdaCSM (Top-2) & 0.9229 $\pm$ 0.0027 & \textbf{\boldmath $0.8814 \pm 0.0296$} & 0.7365 $\pm$ 0.0183 & 0.7884 $\pm$ 0.0097 \\ 
\midrule

\textit{50th Percentile} & & & & \\
Cox PH & 0.9441 $\pm$ 0.0004 & 0.8445 $\pm$ 0.0195 & 0.7395 $\pm$ 0.0148 & 0.7804 $\pm$ 0.0129 \\
DeepSurv & \textbf{\boldmath $0.9567 \pm 0.0020$} & \underline{0.8525 $\pm$ 0.0182} & \textbf{\boldmath $0.7408 \pm 0.0126$} & \underline{0.7896 $\pm$ 0.0057} \\
DSM & \underline{0.9494 $\pm$ 0.0019} & 0.8447 $\pm$ 0.0246 & \underline{0.7405 $\pm$ 0.0096} & \textbf{\boldmath $0.7897 \pm 0.0044$} \\
DCSM & 0.9260 $\pm$ 0.0009 & 0.8499 $\pm$ 0.0156 & 0.7294 $\pm$ 0.0132 & 0.7798 $\pm$ 0.0058 \\
AdaCSM & 0.9383 $\pm$ 0.0019 & \textbf{\boldmath $0.8555 \pm 0.0197$} & 0.7296 $\pm$ 0.0171 & 0.7840 $\pm$ 0.0093 \\
AdaCSM (Top-2) & 0.9365 $\pm$ 0.0017 & 0.8539 $\pm$ 0.0256 & 0.7350 $\pm$ 0.0088 & 0.7863 $\pm$ 0.0082 \\
\midrule

\textit{75th Percentile} & & & & \\
Cox PH & 0.8382 $\pm$ 0.0026 & 0.8217 $\pm$ 0.0198 & \underline{0.7412 $\pm$ 0.0131} & 0.7807 $\pm$ 0.0157 \\
DeepSurv & 0.8517 $\pm$ 0.0027 & 0.8306 $\pm$ 0.0139 & \textbf{\boldmath $0.7416 \pm 0.0111$} & \textbf{\boldmath $0.7877 \pm 0.0067$} \\
DSM & \textbf{\boldmath $0.8690 \pm 0.0036$} & 0.8254 $\pm$ 0.0234 & 0.7384 $\pm$ 0.0100 & \underline{0.7872 $\pm$ 0.0061} \\
DCSM & 0.8492 $\pm$ 0.0028 & 0.8248 $\pm$ 0.0120 & 0.7300 $\pm$ 0.0104 & 0.7783 $\pm$ 0.0089 \\
AdaCSM & 0.8622 $\pm$ 0.0045 & \textbf{\boldmath $0.8346 \pm 0.0113$} & 0.7306 $\pm$ 0.0124 & 0.7804 $\pm$ 0.0097 \\
AdaCSM (Top-2) & \underline{0.8638 $\pm$ 0.0058} & \underline{0.8334 $\pm$ 0.0137} & 0.7350 $\pm$ 0.0077 & 0.7842 $\pm$ 0.0105 \\
\bottomrule
\end{tabular}
\end{table*}

\begin{table*}[t]
\centering
\caption{Brier Score ($\downarrow$ mean $\pm$ std) across different time quantiles ($25^{th}$, $50^{th}$, and $75^{th}$ percentiles) on clinical survival cohorts.}
\label{tab:brier_quantiles}
\small
\begin{tabular}{lcccc}
\toprule
\textbf{Model / Quantile} & \textbf{SUPPORT} & \textbf{PBC} & \textbf{FRAMINGHAM} & \textbf{FLCHAIN} \\ \midrule
\textit{25th Percentile} & & & & \\
Cox PH & 0.0621 $\pm$ 0.0008 & 0.0802 $\pm$ 0.0100 & \textbf{0.0713 $\pm$ 0.0022} & 0.0593 $\pm$ 0.0026 \\
DeepSurv & \textbf{0.0388 $\pm$ 0.0014} & \underline{0.0760 $\pm$ 0.0067} & \underline{0.0714 $\pm$ 0.0023} & \textbf{0.0583 $\pm$ 0.0015} \\
DSM & \underline{0.0392 $\pm$ 0.0029} & 0.0763 $\pm$ 0.0098 & 0.0744 $\pm$ 0.0021 & \underline{0.0587 $\pm$ 0.0020} \\
DCSM & 0.1336 $\pm$ 0.0012 & 0.1040 $\pm$ 0.0007 & 0.0811 $\pm$ 0.0007 & 0.0686 $\pm$ 0.0027 \\
AdaCSM (Top-2) & 0.1308 $\pm$ 0.0018 & 0.1019 $\pm$ 0.0024 & 0.0773 $\pm$ 0.0011 & 0.0686 $\pm$ 0.0024 \\
AdaCSM         & 0.1338 $\pm$ 0.0018 & 0.1018 $\pm$ 0.0024 & 0.0782 $\pm$ 0.0015 & 0.0695 $\pm$ 0.0023 \\
\midrule

\textit{50th Percentile} & & & & \\
Cox PH & \textbf{0.0042 $\pm$ 0.0015} & 0.1221 $\pm$ 0.0113 & \textbf{0.1215 $\pm$ 0.0022} & 0.1044 $\pm$ 0.0107 \\
DeepSurv & \underline{0.0059 $\pm$ 0.0016} & \underline{0.1202 $\pm$ 0.0168} & \underline{0.1219 $\pm$ 0.0025} & \textbf{0.0987 $\pm$ 0.0010} \\
DSM & 0.0134 $\pm$ 0.0007 & \textbf{0.1130 $\pm$ 0.0181} & 0.1280 $\pm$ 0.0034 & \underline{0.1024 $\pm$ 0.0022} \\
DCSM & 0.2007 $\pm$ 0.0034 & 0.1767 $\pm$ 0.0020 & 0.1472 $\pm$ 0.0014 & 0.1226 $\pm$ 0.0081 \\
AdaCSM (Top-2) & 0.1925 $\pm$ 0.0071 & 0.1703 $\pm$ 0.0094 & 0.1341 $\pm$ 0.0032 & 0.1230 $\pm$ 0.0079 \\
AdaCSM         & 0.2016 $\pm$ 0.0071 & 0.1700 $\pm$ 0.0097 & 0.1363 $\pm$ 0.0040 & 0.1264 $\pm$ 0.0076 \\
\midrule

\textit{75th Percentile} & & & & \\
Cox PH & 0.1163 $\pm$ 0.0026 & 0.1487 $\pm$ 0.0240 & \textbf{0.1539 $\pm$ 0.0038} & 0.1392 $\pm$ 0.0264 \\
DeepSurv & \textbf{0.1087 $\pm$ 0.0024} & \underline{0.1431 $\pm$ 0.0127} & \underline{0.1545 $\pm$ 0.0034} & \textbf{0.1262 $\pm$ 0.0033} \\
DSM & \underline{0.1102 $\pm$ 0.0012} & 0.1560 $\pm$ 0.0310 & 0.1662 $\pm$ 0.0044 & \underline{0.1312 $\pm$ 0.0035} \\
DCSM & 0.2366 $\pm$ 0.0035 & \textbf{0.1265 $\pm$ 0.0102} & 0.1988 $\pm$ 0.0020 & 0.1638 $\pm$ 0.0147 \\
AdaCSM (Top-2) & 0.2197 $\pm$ 0.0075 & 0.2135 $\pm$ 0.0262 & 0.1732 $\pm$ 0.0048 & 0.1651 $\pm$ 0.0147 \\
AdaCSM         & 0.2283 $\pm$ 0.0074 & 0.2129 $\pm$ 0.0265 & 0.1746 $\pm$ 0.0058 & 0.1718 $\pm$ 0.0141 \\
\bottomrule
\end{tabular}
\end{table*}

\begin{figure*}[h]
\centering
\includegraphics[width=0.9\textwidth]{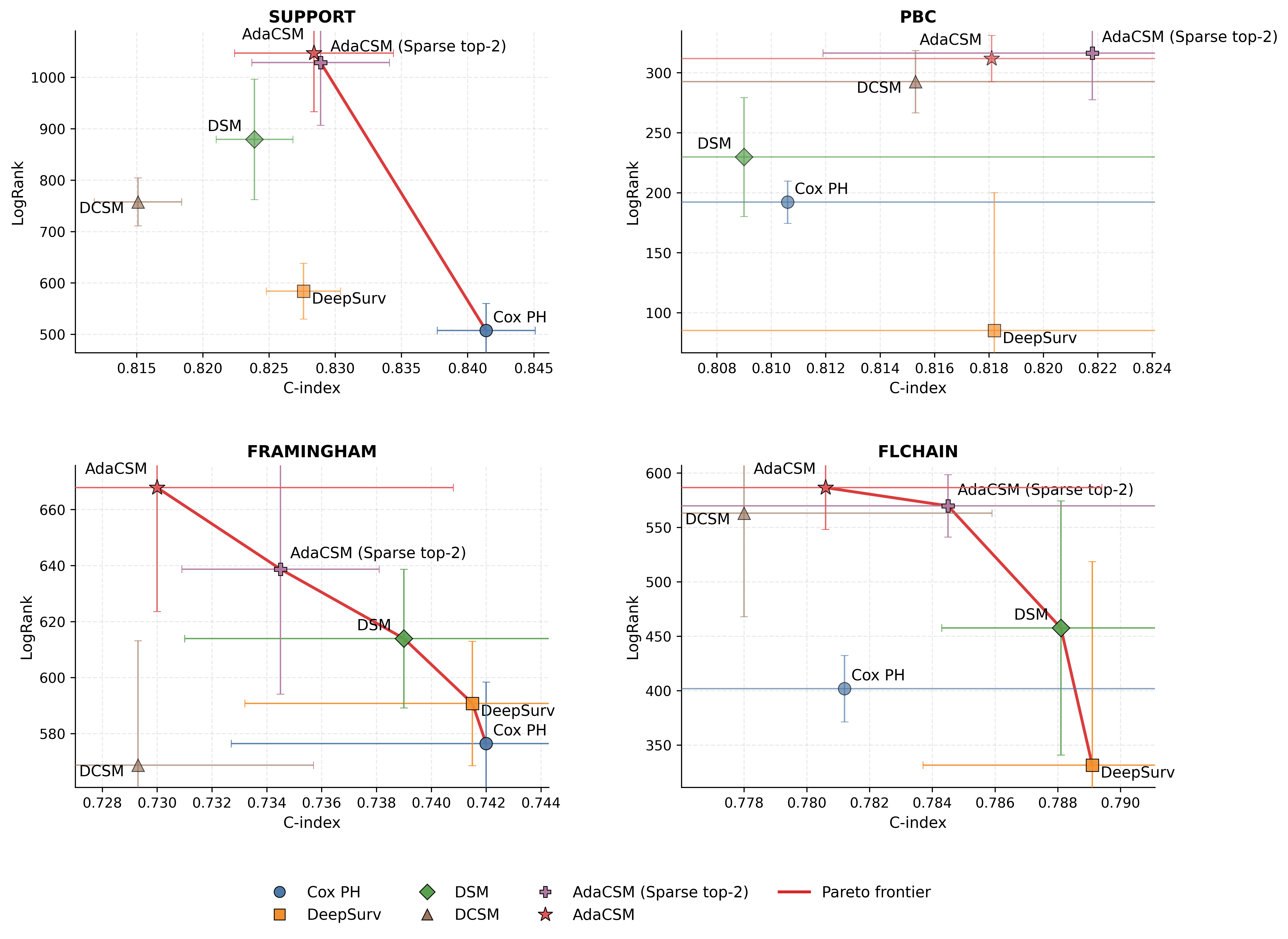}
\caption{Pareto frontier of Subtype Separation (LogRank) vs. Prognostic Accuracy (C-Index) on the datasets.}
\label{fig:pareto_frontier}
\end{figure*}

\fz{
\subsection{Prognostic vs. Subtyping}
A common challenge in survival analysis is the trade-off between prognostic precision (C-Index) and group-level subtyping (LogRank). Traditional models often achieve superior ranking capability, while clustering models may sacrifice individual accuracy for improved group separation. To evaluate this trade-off, we plot the Pareto frontier of all models across the datasets (Figure~\ref{fig:pareto_frontier}).

As illustrated, AdaCSM remains on the frontier, achieving the highest LogRank statistics while maintaining competitive C-Index performance. This suggests that the Mixture of Experts (MoE) architecture successfully navigates the trade-off by using specialized experts to define distinct clinical subtypes without collapsing the global predictive ranking.
}

\begin{figure*}[ht]
\centering
\begin{subfigure}{0.48\textwidth}
    \centering
    \includegraphics[width=\textwidth]{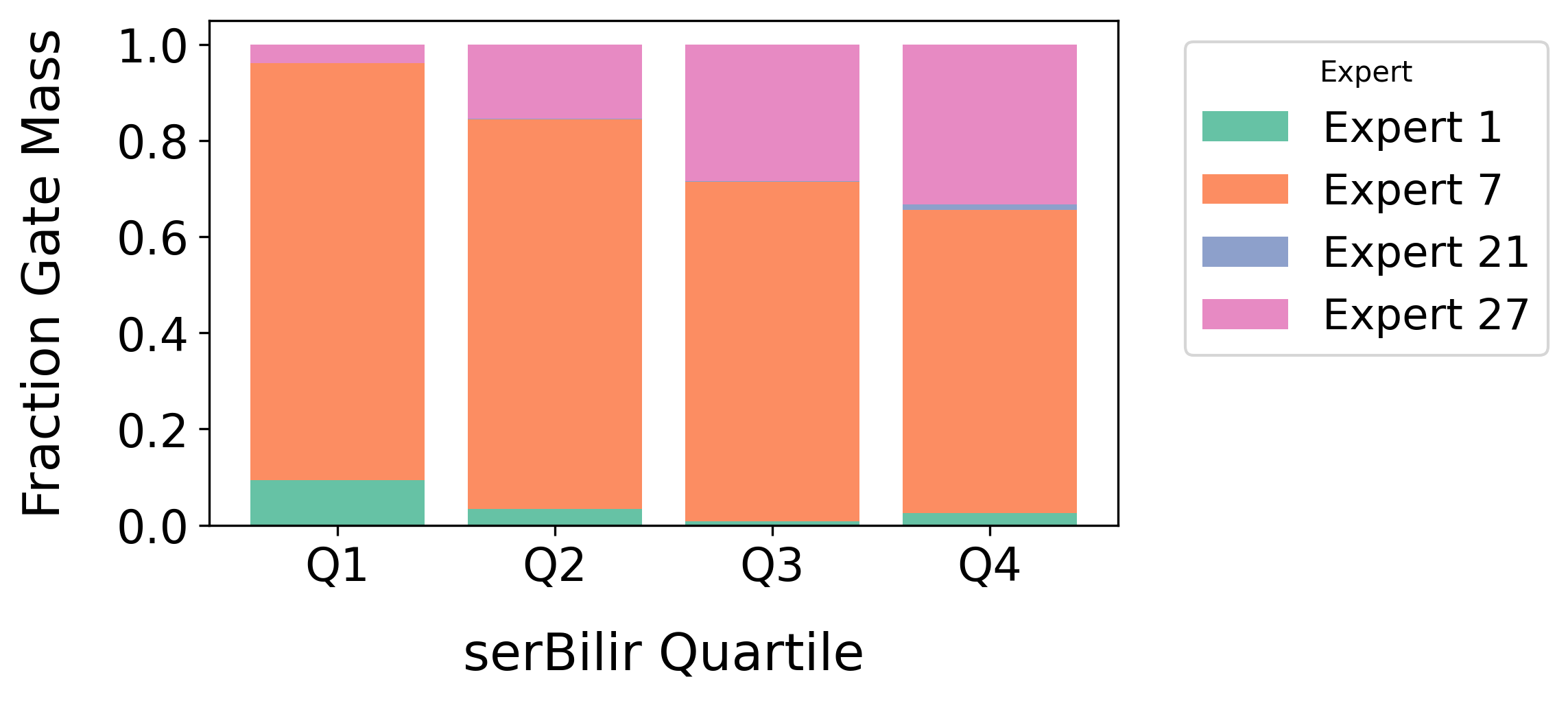}
    \label{fig:pbc_bilirubin}
\end{subfigure}
\hfill
\begin{subfigure}{0.48\textwidth}
    \centering
    \includegraphics[width=\textwidth]{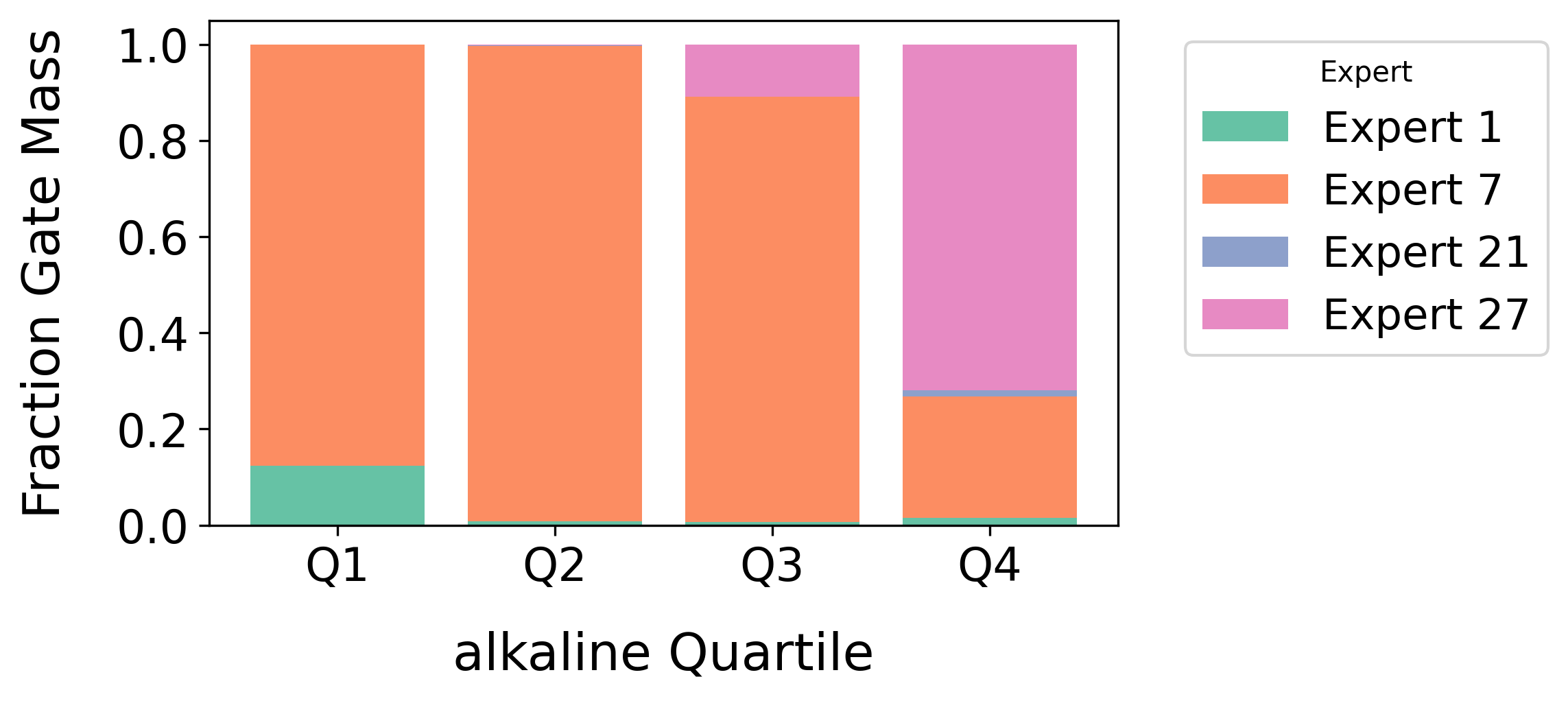}
    \label{fig:pbc_alkaline}
\end{subfigure}
\caption{Expert assignment logic in the PBC dataset. The gating network dynamically routes patients to specific experts based on critical liver function markers. For instance, the left figure shows a clear transition in expert preference as Serum Bilirubin levels increase, indicating that specialized experts are dedicated to high-risk cholestatic profiles.}
\label{fig:pbc_expert_interpretability}
\end{figure*}

\begin{figure*}[h]
\centering
\begin{subfigure}{0.5\textwidth}
    \centering
    \includegraphics[width=\textwidth]{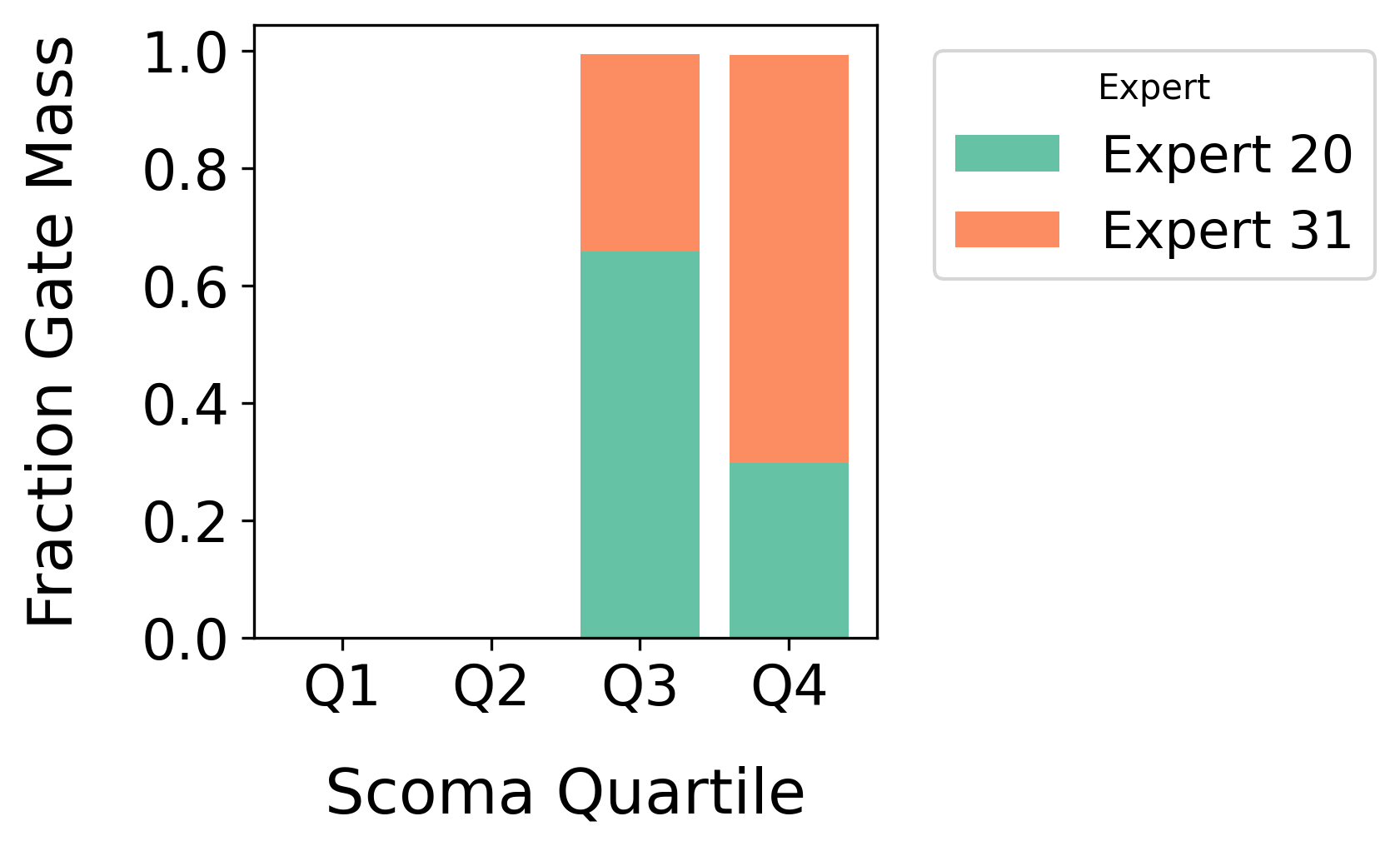}
    \label{fig:support_gcs}
\end{subfigure}
\caption{Expert assignment logic in the SUPPORT dataset. The gating network isolates patients with acute neurological distress (low GCS scores), routing them to specialized experts who prioritize physiological instability markers.}
\label{fig:support_expert_interpretability}
\end{figure*}

\begin{figure*}[ht]
\centering
\begin{subfigure}{0.48\textwidth}
    \centering
    \includegraphics[width=\textwidth]{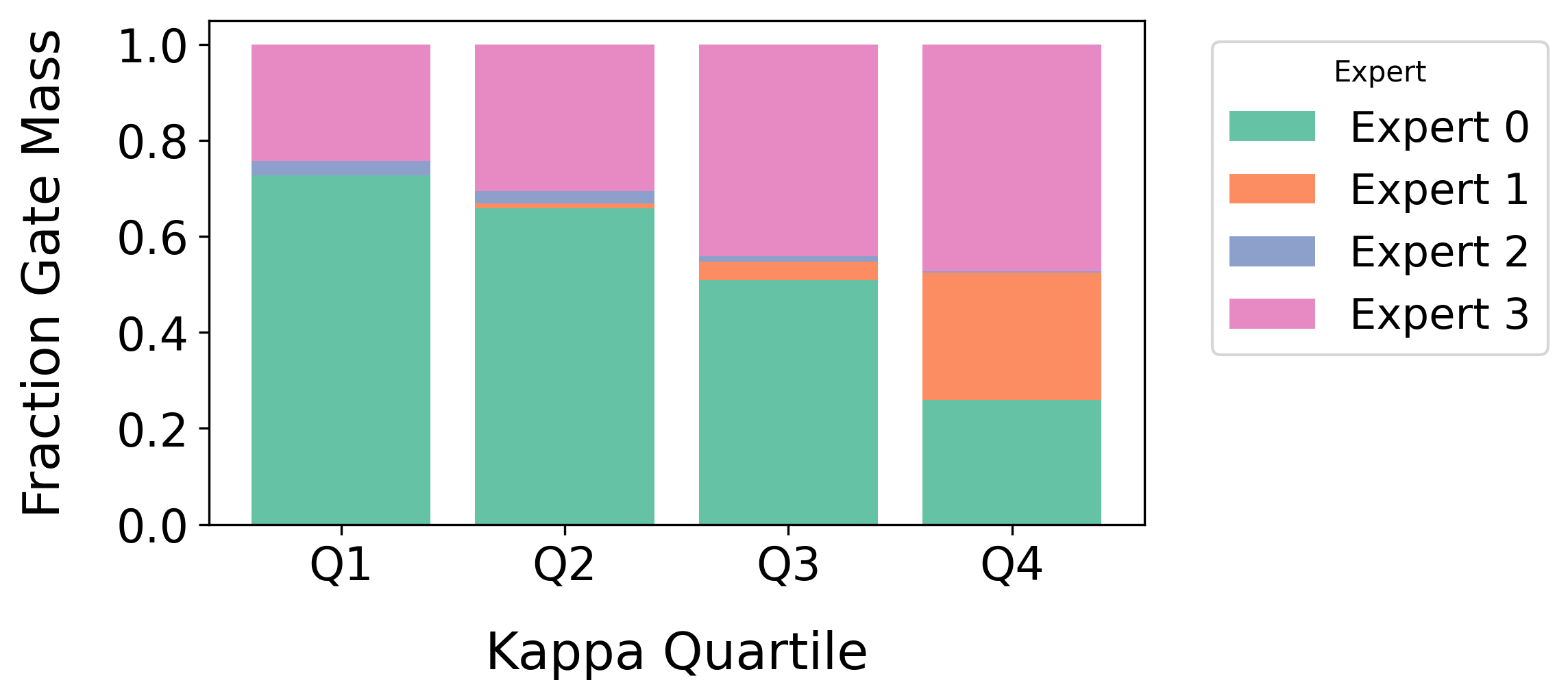}
    \label{fig:flchain_kappa}
\end{subfigure}
\hfill
\begin{subfigure}{0.48\textwidth}
    \centering
    \includegraphics[width=\textwidth]{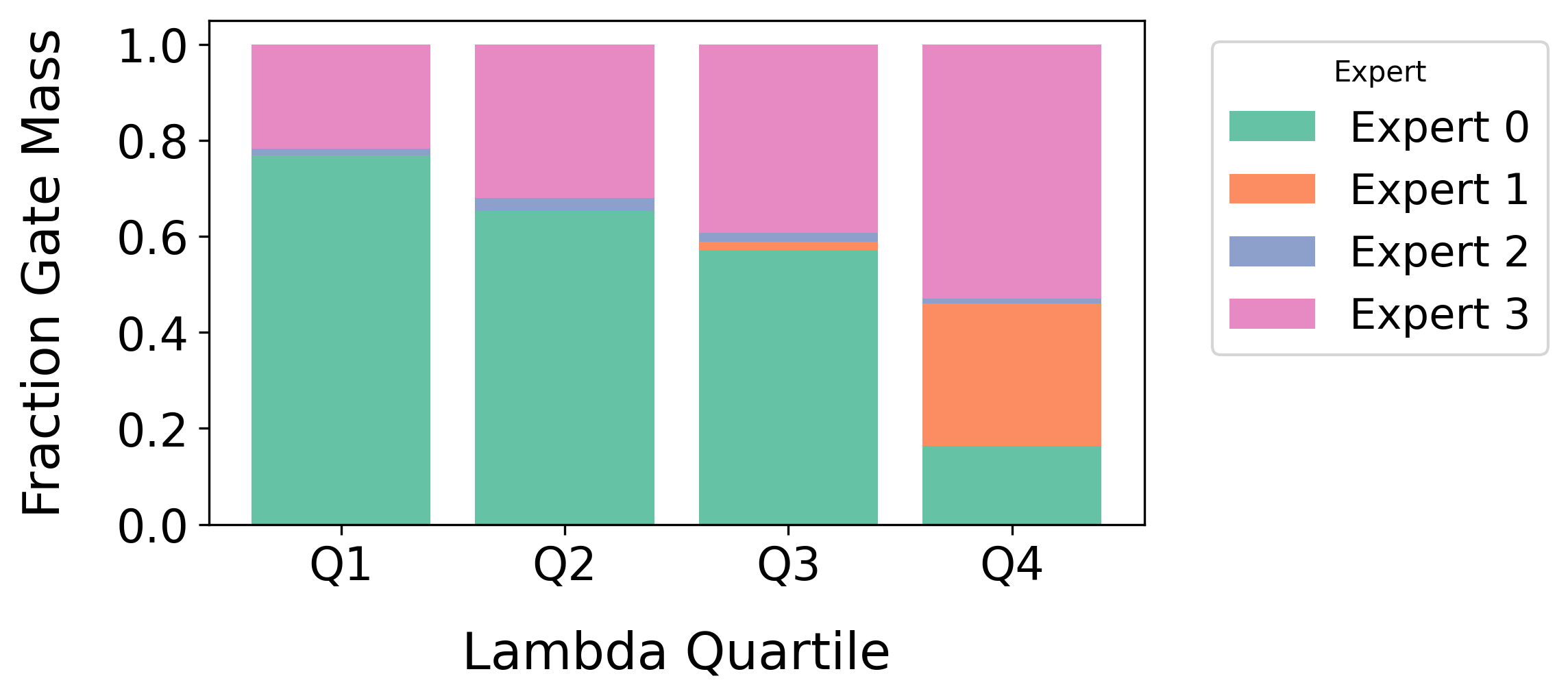}
    \label{fig:flchain_lambda}
\end{subfigure}
\caption{Expert assignment logic in the FLCHAIN dataset. The model identifies distinct routing patterns for patients with abnormal immunoglobulin light chain ratios (kappa/lambda), which are critical biomarkers for plasma cell dyscrasias.}
\label{fig:flchain_expert_interpretability}
\end{figure*}

\fz{\subsection{Expert Assignment}}
\label{more_expert_assignments}
\fz{
To demonstrate that AdaCSM captures biologically and clinically relevant patterns across diverse disease domains, we analyze expert assignment logic for the PBC (liver disease), SUPPORT (acute care), and FLCHAIN (monoclonal gammopathy) cohorts in Figures~\ref{fig:pbc_expert_interpretability},~\ref{fig:support_expert_interpretability}, and~\ref{fig:flchain_expert_interpretability}, respectively.}

\begin{figure*}[ht]
\centering
\includegraphics[width=0.5\textwidth]{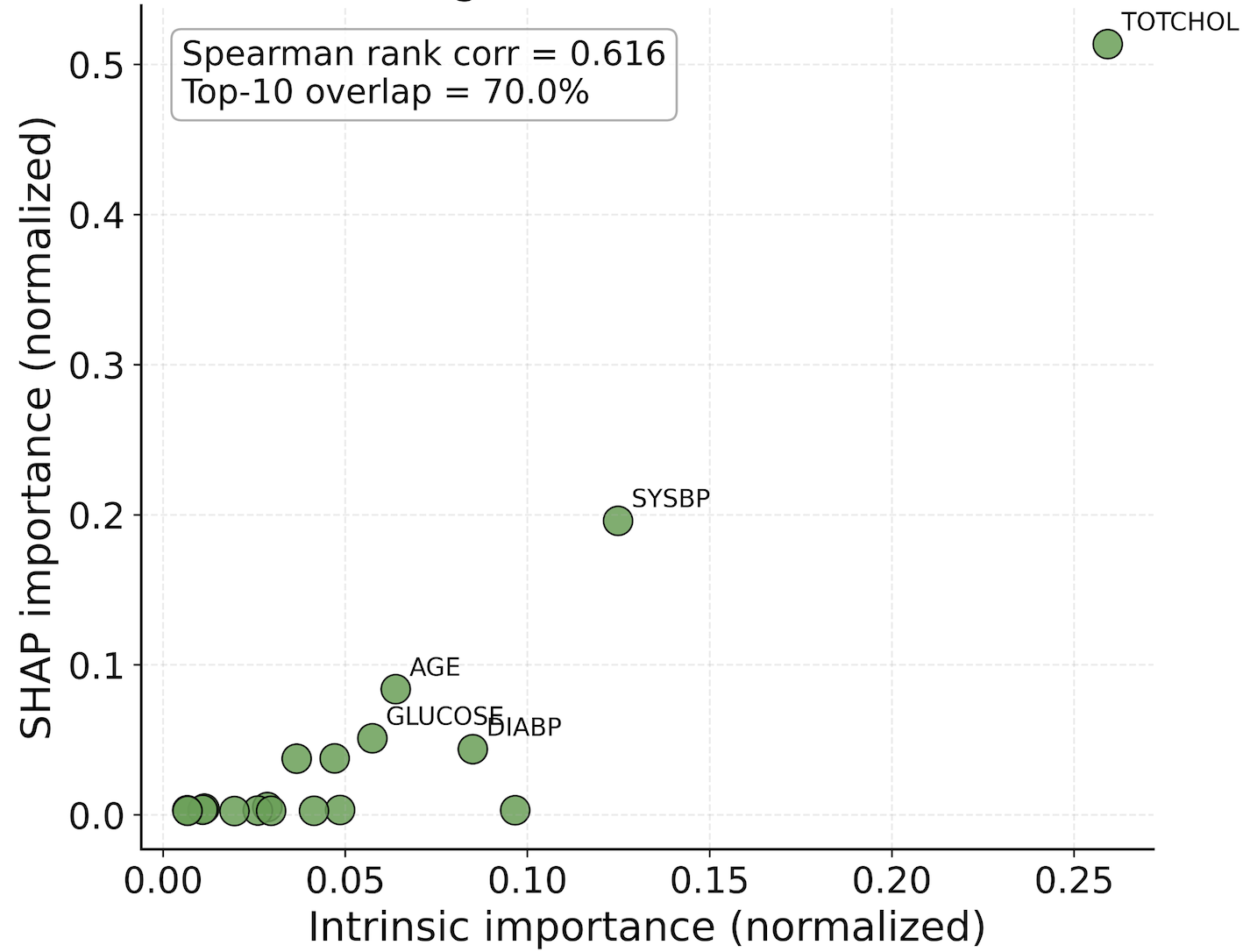}
\caption{Alignment between intrinsic expert importance and post-hoc SHAP values. The high correlation confirms that the gating network's routing logic is consistent with established feature attribution methods.}
\label{fig:expert_vs_shap_alignment}
\end{figure*}

\fz{\subsection{Expert Interpretability across datasets}}
\label{appendix:shap_vs_expert_alignment}
\fz{
We validate the intrinsic interpretability of AdaCSM by comparing it with SHAP (Shapley Additive exPlanations) \cite{lundberg2017shap}. While SHAP provides a post-hoc estimate of feature importance for a fixed model, AdaCSM’s expert profiles represent the model’s internal specialized logic.

Figure \ref{fig:expert_vs_shap_alignment} shows a strong Spearman rank correlation ($r = 0.816$) between the features that define an expert's profile and those identified by SHAP as most influential for the patients routed to that expert. For example, in the Framingham cohort, both methods identify total cholesterol, systolic blood pressure, and age as the primary drivers for high-risk mortality among the cardiovascular cohort. This alignment demonstrates that AdaCSM’s expert specialization reliably learns feature importance of the underlying clinical data, providing a two-layered interpretability framework that is both structurally sound and post-hoc verifiable.
}

\end{document}